\documentclass[sigconf, nonacm]{acmart}

\usepackage{booktabs}
\usepackage{graphicx}
\usepackage{booktabs, graphicx, multirow, makecell, rotating}
\usepackage{geometry}
\usepackage{xcolor}
\usepackage{booktabs,multirow,rotating,graphicx}
\usepackage{colortbl}
\usepackage{subfig}
\usepackage{circledsteps}
\usepackage{xurl}
\usepackage{hyperref}
\usepackage{booktabs}
\usepackage{graphicx}
\usepackage{booktabs, graphicx, multirow, makecell, rotating}
\usepackage{geometry}
\usepackage{xcolor}
\usepackage{booktabs,multirow,rotating,graphicx}
\usepackage{colortbl}
\usepackage{subfig}
\usepackage{circledsteps}
\usepackage{fontawesome5}

\usepackage{pifont}
\newcommand{\cmark}{\ding{51}} 
\newcommand{\xmark}{\ding{55}} 

\definecolor{MyBlue}{RGB}{0, 78, 162}

\makeatletter
\apptocmd{\@mkauthors@iii}{%
  \global\setbox\mktitle@bx=\vbox{%
    \unvbox\mktitle@bx
    \vspace{0.2\baselineskip}
    \centering
    {\small\sffamily
      \begin{tabular}{@{}r@{\hspace{0.6em}}l@{}}
        \textcolor{MyBlue}{\faDatabase} &
        \href{https://outreach.didichuxing.com/opendata/datasets}{%
          \textcolor{MyBlue}{\textbf{Dataset:}}\,
          \textcolor{MyBlue}{outreach.didichuxing.com/opendata/datasets}}\\[0.25em]
        \textcolor{MyBlue}{\faGithub} &
        \href{https://github.com/ACAT-SCUT/ridebench}{%
          \textcolor{MyBlue}{\textbf{Code:}}\,
          \textcolor{MyBlue}{github.com/ACAT-SCUT/ridebench}}
      \end{tabular}\par
    }
    \vspace{0.45\baselineskip}
  }%
}{}{\PackageWarning{main}{Could not add resource links to the title block}}
\makeatother

\begin{document}
  \title[RideBench: A Large-Scale Exogenous-Aware Benchmark for
      Ride-Hailing Time Series Forecasting]{%
    \texorpdfstring{%
      \makebox[\linewidth][c]{%
        \includegraphics[height=0.8cm]{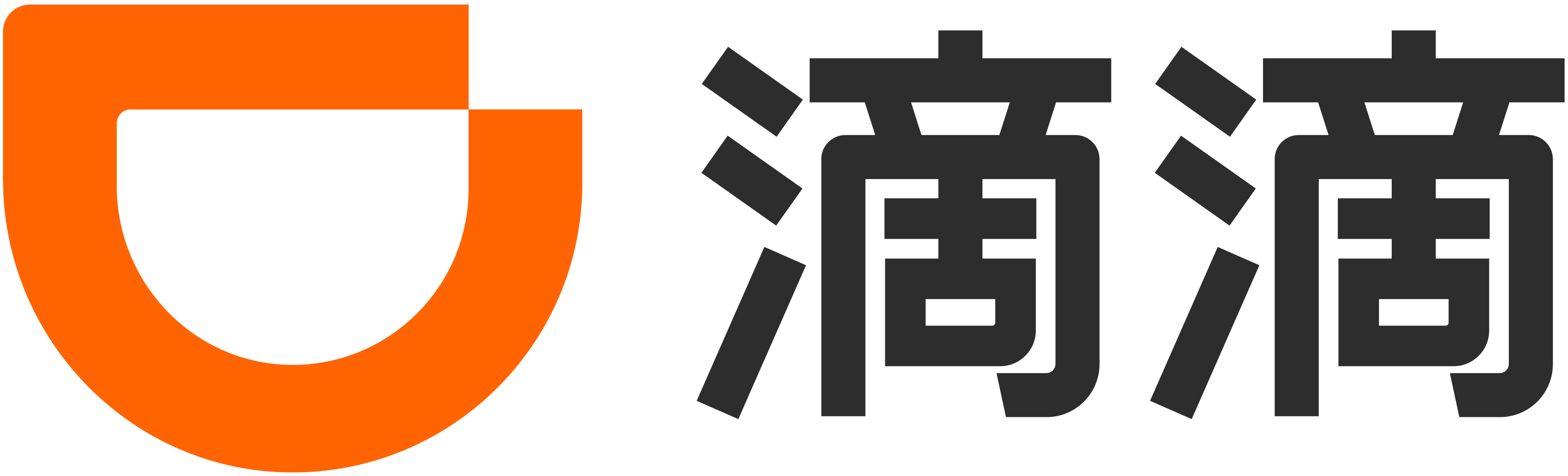}
        \hspace{9cm}
        \includegraphics[height=0.9cm]{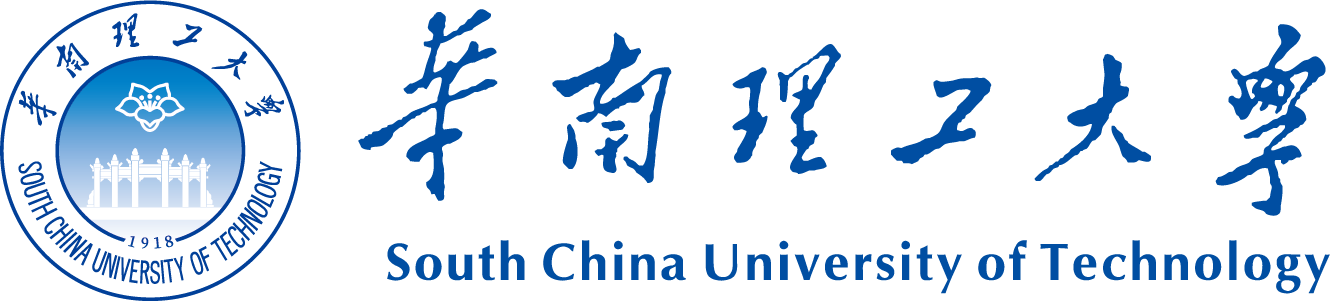}
      }\\[-10pt]
      %
      \rule{0.92\linewidth}{1.6pt}\\[0.1em]
      %
      RideBench: A Large-Scale Exogenous-Aware Benchmark for
      Ride-Hailing Time Series Forecasting\\[-0.3em]
      %
      \rule{0.92\linewidth}{0.6pt}
    }{%
      RideBench: A Large-Scale Exogenous-Aware Benchmark for
      Ride-Hailing Time Series Forecasting
    }%
  }

\settopmatter{authorsperrow=4} 

\author{Shengsheng Lin}
\authornote{Equal contribution.}
\authornote{Work done while at DiDi Chuxing.}
\affiliation{%
  \institution{South China University of Technology, China}
}

\author{Jing Hu}
\authornotemark[1]
\authornotemark[2]
\affiliation{%
  \institution{South China University of Technology, China}
}

\author{Zhengyang Hu}
\authornotemark[2]
\affiliation{%
  \institution{South China University of Technology, China}
}

\author{Jiazheng Sun}
\affiliation{%
  \institution{South China University of Technology, China}
}

\author{Zichun Cao}
\affiliation{%
  \institution{Didi Chuxing, China}
}

\author{Siwei Sun}
\affiliation{%
  \institution{Didi Chuxing, China}
}

\author{Zhichao Zou}
\affiliation{%
  \institution{Didi Chuxing, China}
}

\author{Enyun Yu}
\affiliation{%
  \institution{Didi Chuxing, China}
}

\author{Dongdong Li}
\affiliation{%
  \institution{South China University of Technology, China}
}

\author{Xinyi Hu}
\affiliation{%
  \institution{The University of Hong Kong, China}
}

\author{Weiwei Lin}
\authornote{Corresponding author. Email: linww@scut.edu.cn}
\affiliation{%
  \institution{South China University of Technology, China}
}

\begin{abstract}
We release \textbf{Ride-Hailing}, a large-scale ride-hailing time series dataset synthesized from DiDi's marketplace data across 200 spatial areas.
Ride-Hailing spans four consecutive years at half-hourly granularity and covers three representative exogenous scenarios: \textit{Weather Disturbance}, \textit{Holiday Effect}, and \textit{Large-scale Event Impact}.
Built upon Ride-Hailing, we introduce \textbf{RideBench}, a comprehensive benchmark for exogenous-aware ride-hailing forecasting, covering both regular week-ahead forecasting and long-horizon 8-week-ahead forecasting with up to 2,688 prediction steps.
RideBench evaluates over 30 representative forecasting methods, including endogenous-only models, exogenous-aware models, and time series foundation models.
Our results show that future-known exogenous variables provide clear benefits in regular week-ahead forecasting, especially under weather, holiday, and large-scale event (e.g., major sporting events and concerts) scenarios.
However, current exogenous-aware models still struggle to fully capture disturbance-induced pattern changes under complex external contexts.
For long-horizon forecasting, existing models cannot simultaneously achieve low pointwise errors, accurate broad trends, and reliable near-term forecasts.
These findings reveal a clear mismatch between existing forecasting models and real-world ride-hailing requirements, highlighting the need for models that can better exploit future-known exogenous information, scale across heterogeneous areas, and support long-horizon planning.
By introducing Ride-Hailing and RideBench, we aim to encourage the community to study these practical challenges in real-world ride-hailing forecasting.
\end{abstract}

\maketitle

\section{Introduction}

Ride-hailing platforms have become an essential component of modern urban mobility systems~\cite{spatiotemporal, short}.
Their efficient operation relies heavily on accurate forecasting of marketplace time series, which supports a wide range of platform decisions, including supply-demand balancing, driver incentive planning, resource allocation, and event-specific operations~\cite{didi,dmvst}.
Over the past decade, time series forecasting has made substantial progress~\cite{timesnet, hyperimts, hipatch, cyclenet, informer, stttc, pdf}, driven by deep learning architectures~\cite{transformer,resnet,gnn,mamba,dkgzsl}, large-scale datasets~\cite{quitobench, largest, subseasonalclimateusa}, and standardized benchmarking protocols~\cite{tfb,tslib}.
However, much of this progress has been driven by general-purpose forecasting settings, where benchmarks emphasize endogenous temporal patterns, long-range dependencies, and multivariate correlations more systematically than domain-specific operational requirements.

\begin{figure}
  \centering
  \includegraphics[width=0.95\linewidth]{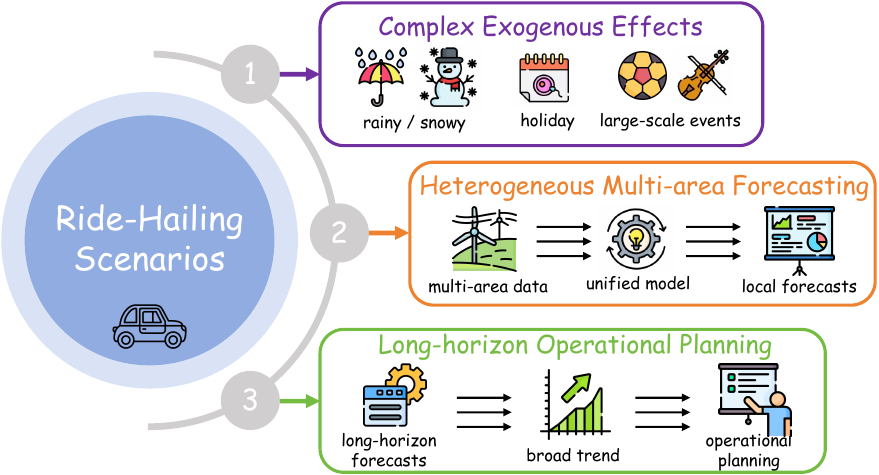}
  \caption{Three key challenges of real-world ride-hailing time series forecasting.}
  \label{fig:ridehailing_overview}
\end{figure}

Ride-hailing forecasting differs from these settings in several important ways.
Unlike many standard forecasting settings~\cite{fincast, ctpatchtst, airformer,spatio}, ride-hailing systems are marketplace-driven, spatially heterogeneous, and highly sensitive to external disturbances.
These properties make forecasts depend not only on endogenous histories, but also on future-known exogenous signals~\cite{dag, gcgnet, tide, tft}, cross-area variation, and longer-term planning needs.
We summarize these requirements into three key challenges, illustrated in Figure~\ref{fig:ridehailing_overview}, that a practical benchmark for this domain should capture.

\textbf{Complex exogenous effects.}
Ride-hailing time series are shaped not only by endogenous temporal dynamics, such as daily and weekly mobility patterns, but also by complex interactions with external factors.
Weather conditions may suddenly alter travel demand and available supply; holidays can reshape commuting routines and leisure travel patterns; and large-scale events (such as major sporting events or concerts) may trigger localized demand surges through crowd movement, road control, or temporary supply-demand imbalance.
Many of these factors are known or can be estimated in advance from external sources, making them valuable for practical forecasting.
Nevertheless, existing benchmarks remain primarily endogenous, providing limited support for future-known exogenous variables, especially categorical external signals such as holidays and large-scale events \cite{monash,gifteval,basicts}.

\textbf{Heterogeneous multi-area forecasting.}
Ride-hailing platforms operate across heterogeneous spatial areas, where time series patterns vary with commuting structure, weather sensitivity, event exposure, and local supply-demand conditions~\cite{didi,dmvst}.
In practice, training a separate model for each area is costly and statistically inefficient, especially when many areas have limited observations under rare external disturbances.
A more practical solution is to train global models across multiple areas, but such models must learn shared temporal and exogenous regularities while preserving area-specific demand patterns.
Existing benchmarks, however, often contain either a limited number of series or lack an evaluation setting designed around representative spatial areas, leaving the scalability and transferability of forecasting models underexplored \cite{fevbench,tslib,probts}.

\textbf{Long-horizon operational planning.}
While short-term and medium-term forecasts, e.g., week-ahead ones, are important for operational dispatching, ride-hailing platforms also require forecasts over much longer horizons for strategic planning, holiday preparation, supply scheduling, and large-scale event management.
In such settings, forecasting is no longer only about minimizing point-wise errors.
A useful model should also capture macro-level temporal trends, including whether the target series will rise or fall and whether the predicted trajectory follows the correct long-term shape.
However, existing benchmarks typically focus on relatively short evaluation horizons and standard error metrics, providing limited evidence about model behavior under long-horizon forecasting \cite{quitobench,time}.

\begin{table}[t]
  \caption{Overview of existing benchmarks and RideBench. }
  \label{tab:benchmark_comparison}
  \centering
  \resizebox{\linewidth}{!}{
  \begin{tabular}{ccccc}
    \toprule
    \textbf{Benchmark} &
    \makecell{\textbf{Future Exo.}\\\textbf{Con./Dis.}} &
    \makecell{\textbf{Scenario}\\\textbf{Eval.}} &
    \makecell{\textbf{Multi-Area}\\\textbf{Modeling}} &
    \makecell{\textbf{Max Eval.}\\\textbf{Horizon}} \\
    \midrule
    Monash~\cite{monash} & \xmark/\xmark & \xmark & \xmark & 168  \\
    TSlib~\cite{tslib}  & \xmark/\xmark & \xmark & \xmark & 720  \\
    BasicTS+~\cite{basicts} & \xmark/\xmark & \xmark & \xmark & 720 \\
    TFB~\cite{tfb} & \cmark/\xmark & \xmark & \xmark & 720 \\
    ProbTS~\cite{probts} & \xmark/\xmark & \xmark & \xmark & 720  \\
    GIFT-Eval~\cite{gifteval} & \xmark/\xmark & \xmark & \xmark & 900  \\
    fev-bench~\cite{fevbench} & \cmark/\cmark & \xmark & \xmark & 168  \\
    QuitoBench~\cite{quitobench} & \xmark/\xmark & \xmark & \xmark & 512  \\
    TIME~\cite{time} & \xmark/\xmark & \xmark & \xmark & 864  \\
    \rowcolor{blue!10} RideBench & \cmark/\cmark & \cmark & \cmark & 2688\\
    \bottomrule
  \end{tabular}
  }%

\end{table}

To bridge these gaps, we introduce \textbf{Ride-Hailing}, a large-scale ride-hailing time series dataset synthesized from DiDi's  marketplace data.
Ride-Hailing contains four consecutive years of half-hourly observations across 200 representative spatial areas, resulting in over 14 million time-stamped records.
Each record includes multiple marketplace target variables together with rich future-known exogenous signals covering three representative scenarios: \textit{Weather Disturbance}, \textit{Holiday Effect}, and \textit{Large-scale Event Impact}.
These scenarios reflect common yet challenging sources of variation in real-world ride-hailing systems, and enable systematic evaluation of exogenous-aware forecasting models.

Building upon Ride-Hailing, we further introduce \textbf{RideBench}, a comprehensive benchmark for exogenous-aware ride-hailing time series forecasting.
It is organized around two complementary forecasting tasks.
The first is \textbf{regular week-ahead forecasting}, which evaluates both overall accuracy and scenario-specific robustness under weather, holiday, and large-scale event impacts.
The second is \textbf{long-horizon 8-week-ahead forecasting}, involving up to 2,688 prediction steps at half-hourly granularity.
Beyond standard pointwise errors, the long-horizon setting further includes day-aggregated trend metrics and first-week accuracy, enabling evaluation of broad trend modeling and near-term precision preservation.

Table~\ref{tab:benchmark_comparison} compares RideBench with existing forecasting benchmarks, highlighting several distinctive advantages.
First, it supports both future-known continuous and discrete exogenous variables, enabling evaluation under realistic exogenously driven scenarios.
Second, it provides explicit scenario-specific evaluation for \textit{Weather Disturbance}, \textit{Holiday Effect}, and \textit{Large-scale Event Impact}, allowing a more fine-grained assessment of model robustness to external perturbations.
Third, it is designed for forecasting across multiple representative spatial areas, making it possible to examine model generalization and transferability under heterogeneous urban contexts.
Finally, RideBench substantially extends the maximum evaluation horizon to 2,688 prediction steps, thereby supporting more challenging long-horizon forecasting studies.

We conduct a comprehensive empirical study on RideBench using over 30 representative forecasting methods, including endogenous-only models~\cite{patchtst,petformer,timemixer,pmdformer,tqnet}, exogenous-aware models~\cite{crosslinear,timexer,xlinear,tide}, and time series foundation models~\cite{chronos2,timers1,timesfm2.5}.
The evaluation examines not only overall forecasting accuracy, but also exogenous utilization, multi-area generalization, and long-horizon temporal modeling.
The results show that current forecasting methods have made clear progress in regular ride-hailing forecasting, but still reveal two key limitations for real-world deployment.

First, although future-known exogenous variables provide clear benefits in regular week-ahead forecasting, current exogenous-aware models still struggle to capture disturbance-induced pattern changes, especially under categorical holidays, sparse large-scale events, and area-dependent exogenous responses.
Second, long-horizon forecasting remains substantially challenging, as existing models cannot simultaneously maintain pointwise accuracy, broad trend quality, and near-term reliability, all of which are important for practical deployment.

These findings show that existing models remain insufficient for the complexity of real-world ride-hailing forecasting.
By providing large-scale data, future-known exogenous variables, and comprehensive evaluation protocols, RideBench serves as a testbed for studying these open challenges.

In summary, our contributions are as follows:
\begin{itemize}
    \item We release \textbf{Ride-Hailing}, a large-scale synthesized ride-hailing time series dataset built upon DiDi's  marketplace data, spanning four years at half-hourly granularity across 200 representative spatial areas, with rich future-known exogenous variables covering weather, holidays, and large-scale events.
    \item We introduce \textbf{RideBench}, an exogenous-aware ride-hailing forecasting benchmark covering both regular week-ahead forecasting and long-horizon 8-week-ahead forecasting, with overall, scenario-specific, multi-area, and trend-level evaluations.
    \item We systematically evaluate over 30 representative forecasting methods under unified protocols, providing a comprehensive empirical comparison for real-world ride-hailing forecasting.
    \item We reveal key limitations of existing models in exploiting future-known exogenous signals and maintaining reliable long-horizon forecasts.
\end{itemize}

\section{Related Work}
\label{sec:related_work}

\textbf{Time series forecasting benchmarks.}
Benchmarks are essential for time series forecasting because they provide unified datasets, data splits, evaluation metrics, and experimental protocols for comparing different models~\cite{quitobench,fevbench}.
Early benchmarks mainly focused on dataset curation and standardized evaluation for common forecasting domains, such as electricity~\cite{hsttn,electricity}, traffic~\cite{spatio, gman, travel}, weather~\cite{airformer, mppn}, and retail demand~\cite{retail,demand}.
Recent efforts further improve benchmarking quality by examining fairness factors in preprocessing, training pipelines, and metric computation, or by extending evaluation to multiple tasks such as forecasting, imputation, classification, and anomaly detection~\cite{tfb,tslib}.
With the rise of time series foundation models, benchmarks such as GIFT-Eval~\cite{gifteval} and TIME~\cite{time} further emphasize zero-shot forecasting and scalable evaluation protocols.

Despite these advances, existing benchmarks still provide limited support for evaluating exogenous-aware forecasting.
Most of them are not designed around marketplace-driven demand, heterogeneous spatial areas, future-known categorical exogenous variables, or long-horizon operational planning.
RideBench complements existing benchmarks by focusing on these practical requirements in a large-scale ride-hailing setting that includes real-world exogenous factors.

\textbf{Time series forecasting models.}
These benchmarks have driven rapid progress in time series forecasting, where methods have evolved from traditional statistical models~\cite{arima,ets,kf,var,knn} to deep learning architectures~\cite{informer,segrnn,dlinear,tqnet,crossgnn,patchtst} and, more recently, foundation models~\cite{fstllm,timellm,timecma,tabpfnts}.
Deep forecasting models can be broadly grouped into channel-independent models~\cite{patchtst,petformer,dlinear,stid,timebase}, multivariate models~\cite{pmdformer,timefilter,crossformer,tqnet,crossgnn}, and exogenous-aware models \cite{timexer,xlinear,tide,cats,dag}.
Channel-independent models improve robustness by modeling each target variable separately, multivariate models capture cross-variable dependencies, and exogenous-aware models incorporate external covariates for context-aware forecasting.
Time series foundation models are pretrained on large-scale cross-domain datasets to improve transferability, generalization, and zero-shot forecasting \cite{chronos2,timevlm,timers1}.

However, existing forecasting models are still not fully aligned with the requirements of ride-hailing forecasting.
Many exogenous-aware models mainly focus on historical or continuous covariates~\cite{crosslinear,timexer,xlinear,dag}, whereas ride-hailing forecasting requires effective use of future-known categorical signals, such as holiday types and large-scale event indicators.
Moreover, models need to capture area-dependent exogenous responses under heterogeneous urban contexts, where the same external factor may induce different demand patterns across areas.
Existing models also tend to emphasize pointwise accuracy~\cite{crosslinear,cyclenet,timexer,gcgnet}, while long-horizon operational planning further requires trend quality and near-term reliability.
These limitations motivate RideBench to jointly evaluate exogenous utilization, multi-area generalization, and long-horizon planning capability.

\begin{figure}
  \centering
  \includegraphics[width=\linewidth]{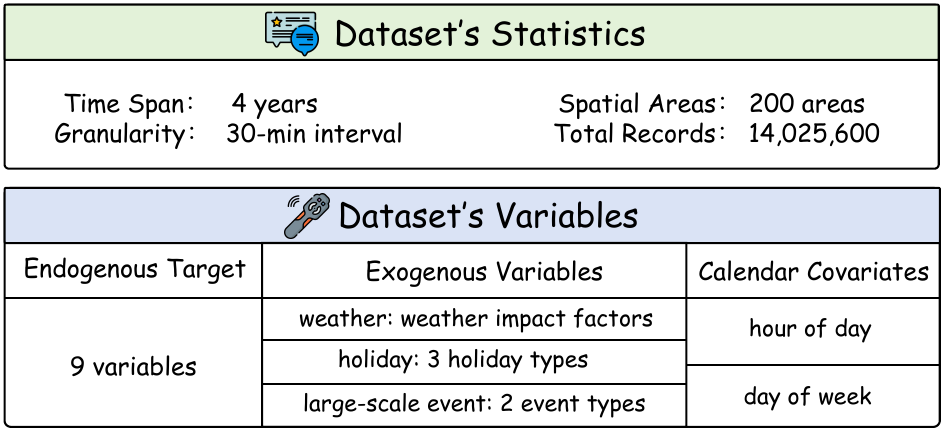}
  \caption{Statistics and variable information of Ride-Hailing.}
  \label{fig:ride_hailing}
\end{figure}

\begin{figure*}
  \centering
  \includegraphics[width=0.95\linewidth]{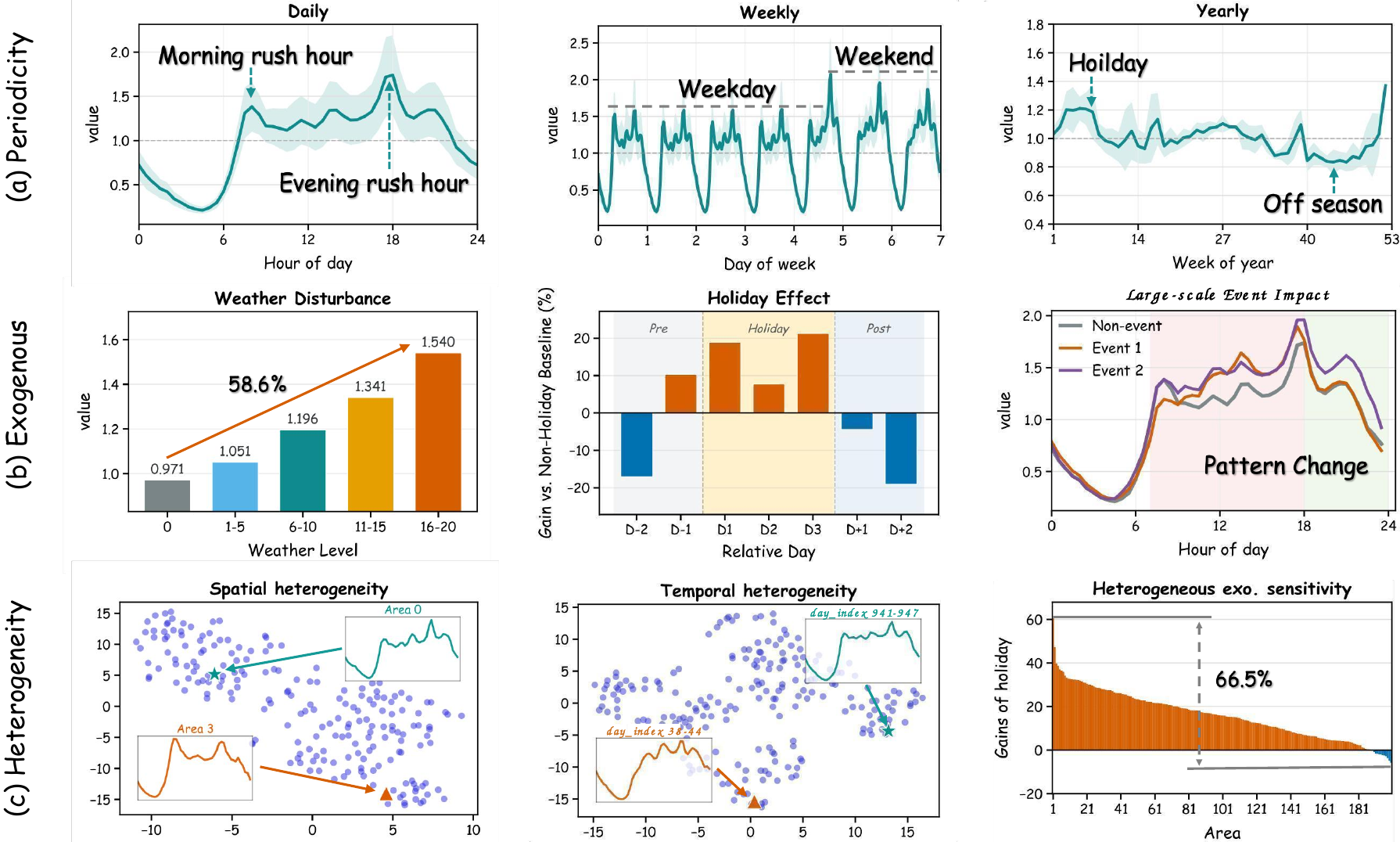}
  \caption{Illustration of the key characteristics of Ride-Hailing. (a) Multi-scale periodicity at daily, weekly, and yearly levels. (b) Exogenous effects under weather disturbance, holiday effect, and large-scale event impact. (c) Heterogeneous patterns across different areas, time periods and exogenous variables, where t-SNE~\cite{tsne} is used to visualize spatial and temporal heterogeneity.}
  \label{fig:dataset}
\end{figure*}

\section{Dataset}
\label{sec:dataset}
\subsection{Dataset Construction}

\textbf{Data sources and scope.}
As shown in Figure \ref{fig:ride_hailing}, Ride-Hailing is synthesized from DiDi's  ride-hailing marketplace data collected across 200 representative spatial areas.
It covers a continuous four-year period at a 30-minute granularity, yielding 14,025,600 time-stamped observations in total.
Each observation is indexed by region identifier and timestamp, and includes both historical marketplace variables and exogenous contextual features.

\textbf{Privacy-preserving data processing.}
Ride-Hailing is built upon  operational records from DiDi's ride-hailing marketplace and is released as a synthesized dataset to address privacy protection and business security requirements.
To this end, we apply a series of anonymization and transformation procedures to avoid exposing sensitive business information while retaining the core temporal signals needed for forecasting research.
These procedures include, but are not limited to, anonymizing area and variable identifiers, processing endogenous target variables, aggregating weather disturbances, and representing large-scale events with coarse-grained impact categories rather than raw event descriptions.
In this way, Ride-Hailing preserves, to the greatest extent possible, the temporal characteristics, exogenous relationships, and cross-area structures of real-world ride-hailing systems, supporting academic research without disclosing sensitive operational details.

\textbf{Endogenous forecasting targets.}
Ride-Hailing provides 9 platform-level endogenous variables, anonymized as endo-1 to endo-9. 
These variables may interact through marketplace mechanisms such as disturbance, substitution, and co-movement, and serve as the forecasting targets in RideBench.

\textbf{Exogenous variables.}
To support exogenous-aware forecasting, Ride-Hailing incorporates future-known variables covering three scenarios: \textit{Weather Disturbance}, \textit{Holiday Effect}, and \textit{Large-scale Event Impact}.
Weather is represented by an impact factor, holidays by categorical variables covering public holidays, traditional festivals, and Western festivals, and large-scale events by two event-impact variables capturing different mobility changes.
Calendar covariates, including hour-of-day and day-of-week indicators, are also included to model regular temporal patterns.

\subsection{Dataset Characteristics}

As marketplace-driven urban mobility data, ride-hailing time series exhibit strong periodicity, sensitivity to exogenous disturbances, and substantial heterogeneity across areas, time periods, and exogenous conditions.

\subsubsection{Multi-scale Periodicity}
Ride-hailing time series exhibit clear multi-scale periodic structures at daily, weekly, and yearly scales, as shown in Figure~\ref{fig:dataset}(a). 
These coupled periodic patterns reflect the regular rhythms of urban activities and human mobility.

\textbf{Daily periodicity.}
At the daily scale, ride-hailing demand changes substantially within a day, following human mobility routines such as commuting and daily activities. 
Demand usually remains low during late-night hours when city activities are limited, and reaches major peaks during morning or evening rush hours due to intensive commuting demand.

\textbf{Weekly periodicity.}
At the weekly scale, demand patterns differ markedly between weekdays and weekends. 
Weekdays typically exhibit commuting-related morning peaks, whereas weekends show weaker commuting regularity and more leisure-oriented demand. 
Friday, although still a weekday, often presents particularly strong evening demand because it is close to the weekend. 

\textbf{Yearly periodicity.}
At the yearly scale, ride-hailing demand also shows evident seasonal variation, although the pattern becomes smoother after weekly aggregation. 
Seasonal changes, public holidays, vacations, and large-scale travel periods can alter the overall demand level and reshape long-term temporal trends.

These multi-scale patterns require models to capture both short-term routines and long-term seasonal evolution, which is especially important for the long-horizon 8-week-ahead forecasting task.

\subsubsection{Exogenous Effects}

Ride-Hailing contains rich future-known exogenous variables, allowing us to examine how external factors are associated with changes in ride-hailing dynamics.
Figure~\ref{fig:dataset}(b) presents representative patterns under weather disturbance, holiday periods, and large-scale events.

\textbf{Weather Disturbance.}
Weather changes can reshape ride-hailing dynamics by altering mobility choices and transport convenience.
As weather disturbance becomes stronger, the average endogenous value increases substantially, indicating higher dependence on ride-hailing services under adverse weather.
Beyond such level-driven effects, extreme weather may also introduce delayed impacts, making the forecasting problem more challenging.

\textbf{Holiday Effect.}
Holidays introduce structured but non-uniform deviations from regular patterns, as ride-hailing demand does not simply increase or decrease throughout the holiday period.
Instead, demand often exhibits stage-dependent changes: distinct demand shifts may emerge before the official holiday starts, fluctuate across in-holiday stages, and rebound during post-holiday return flows.
Moreover, days farther before or after the holiday may experience reduced regular demand, potentially because mobility demand is shifted toward or absorbed by the holiday period.
These stage-dependent effects make holidays more challenging than simple binary calendar indicators.

\textbf{Large-scale Event Impact.}
Large-scale events can cause localized and time-dependent pattern changes. Different event types may also affect demand in different ways. 
In Ride-Hailing, large-scale events are collected and represented using two coarse-grained event-impact categories: one mainly changes daytime mobility, while the other may affect both daytime and evening demand. 
This provides a challenging test of whether models can exploit event-type information and learn event-specific temporal responses.

\subsubsection{Heterogeneity}
Beyond periodicity and exogenous effects, Ride-Hailing also exhibits strong heterogeneity, as shown in Figure~\ref{fig:dataset}(c).
This is particularly important in RideBench, where a global model is trained across multiple areas but must generate reliable forecasts for each individual area.
Therefore, effective models need to learn shared regularities while preserving area-specific temporal patterns.

\textbf{Spatial heterogeneity.}
Different areas have distinct mobility structures.
Some areas show clear morning and evening commuting peaks, while others are dominated by evening activities.
Such differences may reflect land use, commuting function, commercial activity, residential density, and local supply-demand conditions.

\textbf{Temporal heterogeneity.}
The temporal pattern of the same system can also evolve over years.
For instance, the daily curve of an area may change from a multi-peak pattern to a more concentrated evening-peak pattern.
Models need to adapt to latent temporal shifts caused by urban development, behavioral changes, and evolving platform operations.

\textbf{Heterogeneous exogenous sensitivity.}
The impact of the same exogenous factor can also vary across areas. 
For example, although holidays generally stimulate ride-hailing demand, the magnitude of the response differs substantially across areas. 
Some areas may even show negative changes, suggesting that holiday mobility can shift demand away from certain areas or interact with other area-specific factors. 
Similarly, weather disturbances and large-scale events may produce markedly different effects across areas.

In summary, Ride-Hailing provides a realistic testbed where models must jointly handle endogenous temporal dynamics, future-known exogenous variables, and heterogeneous multi-area patterns. 
These characteristics make Ride-Hailing a challenging benchmark foundation for evaluating whether forecasting models can generalize under  marketplace dynamics.

\section{Benchmark}
\label{sec:benchmark}

\begin{figure*}
  \centering
  \includegraphics[width=0.9\linewidth]{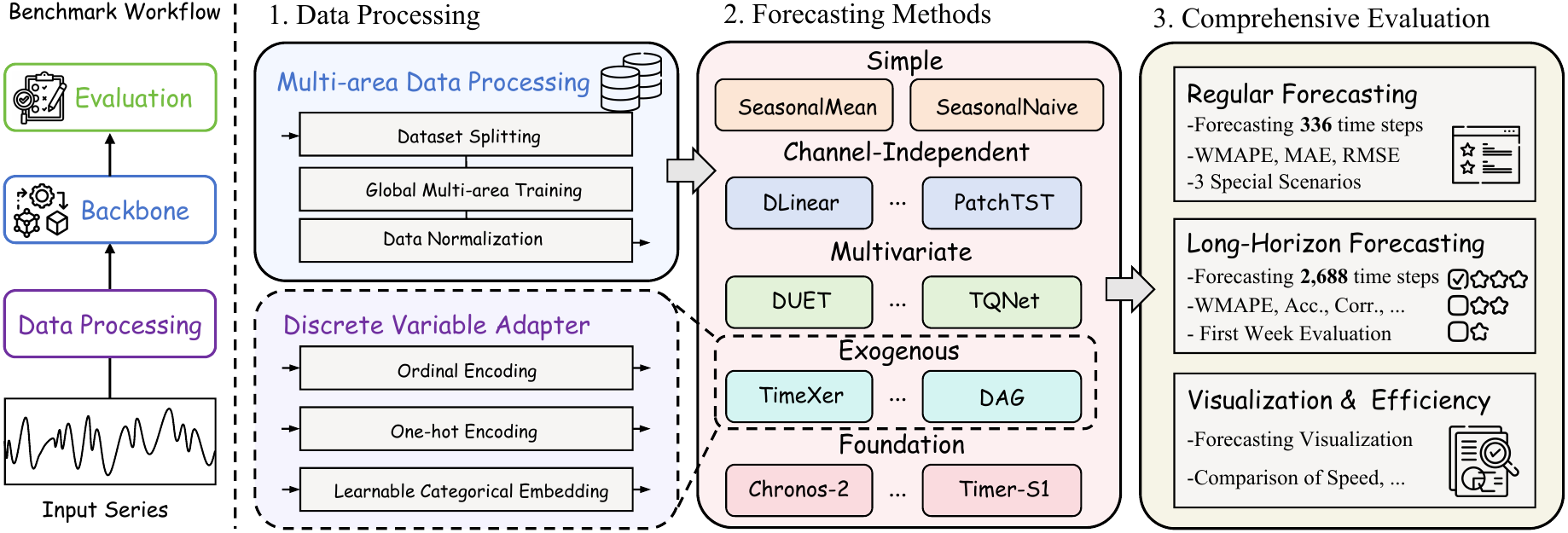}
  \caption{Overall architecture of RideBench. The left illustrates the general benchmark workflow, and the right demonstrates the detailed implementation details.}
  \label{fig:benchmark}
\end{figure*}

Built upon Ride-Hailing, we introduce \textbf{RideBench}, a unified benchmark for evaluating exogenous-aware ride-hailing time series forecasting.
As shown in Figure~\ref{fig:benchmark}, RideBench consists of three core components: standardized data processing, representative forecasting methods, and comprehensive evaluation protocols.

\subsection{Data Processing}

\textbf{Dataset splitting.}
RideBench first constructs forecasting samples using a sliding-window strategy and then assigns them to different splits according to fixed chronological boundaries.
Specifically, the first three years of observations are used for training, the following six months for validation, and the final six months for testing.
For each generated sample, the split assignment is determined by its prediction window: a sample is included in a split only when its full target range falls within the corresponding time period.
The input window is allowed to extend to earlier observations, so that validation and test samples can use realistic historical context without leaking future target values across splits.

\textbf{Global multi-area training.}
For each temporal window, every area is treated as an independent forecasting sample, and samples from all 200 areas are pooled to train a shared global model.
This protocol better reflects practical deployment, where maintaining a separate model for every area is costly and less scalable.
It also poses a stronger modeling challenge, since models must learn transferable temporal structures while preserving area-specific characteristics under substantial spatial heterogeneity.

\textbf{Normalization.}
RideBench standardizes endogenous variables and continuous exogenous variables using statistics fitted only on the training time range, while keeping discrete exogenous variables as integer identifiers.
During evaluation, the predicted endogenous variables are inverse-transformed back to the original scale using the same training statistics, so that forecasting metrics are computed on realistic value scales.

\subsection{Forecasting Methods}
RideBench comprehensively evaluates more than 30 representative forecasting methods spanning five major methodological paradigms.

\textbf{Simple Statistical Baselines} include SeasonalNaive~\cite{seasonalnaive} and SeasonalMean. 
SeasonalNaive directly repeats values from the previous seasonal cycle, while SeasonalMean forecasts each future step using the average of 
historical observations at the same seasonal position within the input window. These two methods capture recurring daily and weekly patterns
without model training, providing simple references for assessing the gains of learned forecasting models.

\textbf{Channel-Independent Forecasting Models} include DLinear~\cite{dlinear}, PatchTST~\cite{patchtst}, PETformer~\cite{petformer}, STID~\cite{stid}, RMLP~\cite{rmlp}, SegRNN~\cite{segrnn}, SparseTSF~\cite{sparsetsf}, PhaseFormer~\cite{phaseformer}, TimeBase~\cite{timebase}, and TimeMixer~\cite{timemixer}.
These models are evaluated on all endogenous forecasting targets, but adopt a channel-independent modeling strategy, where each endogenous variable is modeled based on its own temporal evolution.
This group allows us to examine how much predictive power can be obtained from individual temporal dynamics alone, without explicitly modeling dependencies among endogenous variables in the multi-target ride-hailing forecasting task.

\textbf{Multivariate Forecasting Models} include Crossformer~\cite{crossformer}, CrossGNN~\cite{crossgnn}, DUET~\cite{duet}, Leddam~\cite{leddam}, ModernTCN~\cite{moderntcn}, PMDformer~\cite{pmdformer}, SOFTS~\cite{softs}, TimeFilter~\cite{timefilter}, TQNet~\cite{tqnet}, and iTransformer~\cite{itransformer}.
These models also forecast all endogenous targets, but further capture joint structures among variables through cross-channel dependencies or shared multivariate representations.
This group allows us to evaluate whether interactions among different marketplace signals provide additional benefits beyond independent temporal modeling.

\textbf{Exogenous-Aware Forecasting Models} include CATS~\cite{cats}, CrossLinear~\cite{crosslinear}, DAG~\cite{dag}, TiDE~\cite{tide}, TimeXer~\cite{timexer}, and XLinear~\cite{xlinear}.
These models are closely aligned with the ride-hailing forecasting scenario considered in RideBench, where future-known external factors such as weather, holidays, and large-scale events may substantially affect marketplace dynamics.
However, most existing exogenous-aware models are primarily designed for continuous covariates, whereas RideBench contains both continuous and discrete exogenous variables.
To make these methods applicable to our setting, we adapt discrete exogenous variables using three encoding strategies: ordinal encoding, one-hot encoding, and learnable categorical embeddings.
For models that do not originally expose future covariates to the prediction horizon, we further modify their input interfaces so that future-known exogenous variables can be incorporated during forecasting.

\textbf{Time Series Foundation Models} include Chronos-2~\cite{chronos2}, Moirai 2.0~\cite{moirai2.0}, TimesFM 2.5~\cite{timesfm2.5}, and Timer-S1~\cite{timers1}.
These models can be directly applied to downstream forecasting tasks in a zero-shot manner without task-specific training.
Most of them follow a univariate forecasting paradigm, while Chronos-2 is the only one that explicitly supports multivariate endogenous interactions and future-known covariates, including both continuous and discrete exogenous variables.
We include these models in RideBench to examine how current time series foundation models perform in realistic, fine-grained, and exogenous-driven ride-hailing scenarios.

\subsection{Comprehensive Evaluation}
\subsubsection{Forecasting Settings}

RideBench provides two forecasting settings at half-hourly granularity.
The first is \textbf{week-ahead forecasting}, where models predict the next 336 steps, corresponding to one week, using 1,344 historical steps, corresponding to four weeks.
This setting matches  operational needs~\cite{didi} such as demand monitoring, supply-demand balancing, resource allocation, and event-specific operations under external disturbances.

The second is \textbf{8-week-ahead forecasting}, where models predict the next 2,688 steps, corresponding to eight weeks, using 4,704 historical steps, corresponding to fourteen weeks.
This setting substantially extends the forecasting horizon from short-term fluctuation modeling to long-term temporal evolution modeling.
It is designed to examine whether forecasting models can capture macro-level demand trends over multiple weeks, which is critical for strategic planning, holiday operations, and large-scale resource scheduling.

\subsubsection{Evaluation Protocol}

RideBench evaluates forecasting models from three complementary perspectives: overall forecasting accuracy, scenario-specific robustness, and long-horizon trend quality.

\textbf{Overall forecasting accuracy.}
We evaluate point forecasting accuracy using three standard metrics: Weighted Mean Absolute Percentage Error (WMAPE), Mean Absolute Error (MAE), and Root Mean Squared Error (RMSE).
WMAPE measures scale-normalized absolute error, MAE measures average absolute error, and RMSE emphasizes large deviations.
Formally,

\begin{equation}
\operatorname{WMAPE}
=
\frac{\sum_{i=1}^{n}|y_i-\hat{y}_i|}
{\sum_{i=1}^{n}|y_i|},
\end{equation}
\begin{equation}
\operatorname{MAE}
=
\frac{1}{n}\sum_{i=1}^{n}|y_i-\hat{y}_i|,
\end{equation}
\begin{equation}
\operatorname{RMSE}
=
\sqrt{
\frac{1}{n}\sum_{i=1}^{n}(y_i-\hat{y}_i)^2
},
\end{equation}
where $y_i$ and $\hat{y}_i$ denote the ground-truth and predicted values of the $i$-th evaluated point, and $n$ is the total number of evaluated points.

\textbf{Scenario-specific evaluation.}
To examine whether models can handle external disturbances, RideBench further reports forecasting performance under three representative scenarios: \textit{Weather Disturbance}, \textit{Holiday Effect}, and \textit{Large-scale Event Impact}.
For each scenario, we use the corresponding future-known exogenous indicators to select test periods affected by weather disturbances, holidays, or large-scale events, and compute metrics on these scenario-specific subsets.
This protocol tests whether a model can exploit future-known external information, rather than only fitting regular daily or weekly periodicity.

\textbf{Long-horizon trend evaluation.}
For the 8-week-ahead setting, pointwise errors alone are insufficient to assess whether a model captures the correct long-term trajectory.
Therefore, RideBench additionally evaluates day-level trend quality using Directional Accuracy of Day (Acc.) and Correlation of Day (Corr.).

We first aggregate half-hourly predictions into daily values by averaging every 48 time steps.
Let $Y_i$ and $\hat{Y}_i$ denote the ground-truth and predicted daily values on day $i$, respectively.
Acc. is defined as
\begin{equation}
\operatorname{Acc.}
=
\frac{1}{|\Omega|}
\sum_{i\in\Omega}
\mathbb{I}
\left[
\operatorname{sign}(Y_i-Y_{i-1})
=
\operatorname{sign}(\hat{Y}_i-\hat{Y}_{i-1})
\right],
\end{equation}
where $\Omega=\{i: |Y_i-Y_{i-1}|>10^{-6}\}$ denotes the set of days with non-negligible ground-truth changes.
This metric measures whether the model predicts the correct direction of day-to-day demand changes.
Corr. is defined as
\begin{equation}
\operatorname{Corr.}
=
\frac{
\sum_{i=1}^{m}(Y_i-\mu)(\hat{Y}_i-\hat{\mu})
}{
\sqrt{\sum_{i=1}^{m}(Y_i-\mu)^2}
\sqrt{\sum_{i=1}^{m}(\hat{Y}_i-\hat{\mu})^2}
},
\end{equation}
where $m$ is the number of evaluated days, and $\mu$ and $\hat{\mu}$ are the means of the ground-truth and predicted daily sequences, respectively.
Corr. measures whether the predicted daily trajectory follows the overall shape of the ground truth.
Together, Acc. and Corr. provide a macro-level view of long-horizon forecasting quality beyond point-wise errors.

\begin{table*}[t]
\centering
\caption{Regular week-ahead forecasting performance under the overall evaluation and the three exogenous scenarios. Darker shading indicates better performance. 
Average rank is obtained via arithmetic mean of ranks across all evaluation metrics. }
\resizebox{0.88\textwidth}{!}{%
\begin{tabular}{c|c|ccc|ccc|ccc|ccc|c}
\toprule
& Scenarios & \multicolumn{3}{c|}{Overall} & \multicolumn{3}{c|}{Weather Disturbance} & \multicolumn{3}{c|}{Holiday Effect} & \multicolumn{3}{c|}{Large-scale Event Impact} & Rank \\
\cmidrule(lr){3-5} \cmidrule(lr){6-8} \cmidrule(lr){9-11} \cmidrule(lr){12-14}
\multicolumn{1}{c|}{} & \multicolumn{1}{c|}{Metrics} & \multicolumn{1}{c}{WMAPE} & \multicolumn{1}{c}{MAE} & \multicolumn{1}{c|}{RMSE} & \multicolumn{1}{c}{WMAPE} & \multicolumn{1}{c}{MAE} & \multicolumn{1}{c|}{RMSE} & \multicolumn{1}{c}{WMAPE} & \multicolumn{1}{c}{MAE} & \multicolumn{1}{c|}{RMSE} & \multicolumn{1}{c}{WMAPE} & \multicolumn{1}{c}{MAE} & \multicolumn{1}{c}{RMSE} & \multicolumn{1}{|c}{Average} \\ \midrule
 & SeasonalMean & 0.1285 & 4.916 & 14.37 & 0.1509 & 6.311 & 20.90 & 0.1701 & 7.052 & 20.85 & 0.1224 & 5.395 & \cellcolor[HTML]{E9E9FE}16.32 & 29 \\
\multirow{-2}{*}{\rotatebox[origin=c]{90}{Sim.}} & SeasonalNaive & 0.1490 & 5.693 & 17.75 & 0.1755 & 7.338 & 24.41 & 0.1981 & 8.203 & 24.57 & 0.1422 & 6.261 & 19.10 & 32 \\
\midrule  
& PatchTST & \cellcolor[HTML]{E6E6FE}0.1079 & \cellcolor[HTML]{E6E6FE}4.132 & 13.38 & 0.1391 & 5.821 & 20.84 & \cellcolor[HTML]{F3F3FF}0.1507 & \cellcolor[HTML]{F3F3FF}6.249 & 20.37 & \cellcolor[HTML]{CFCFFE}0.1130 & \cellcolor[HTML]{DBDBFE}4.983 & 16.56 & 10 \\
 & PETformer & \cellcolor[HTML]{DBDBFE}0.1078 & \cellcolor[HTML]{DBDBFE}4.124 & \cellcolor[HTML]{F9F9FF}13.35 & 0.1396 & 5.839 & 20.86 & 0.1520 & 6.303 & 20.42 & \cellcolor[HTML]{CFCFFE}0.1130 & \cellcolor[HTML]{CFCFFE}4.982 & 16.48 & 11 \\
 & TimeMixer & 0.1113 & 4.261 & 13.47 & 0.1411 & 5.902 & 20.90 & 0.1548 & 6.420 & 20.42 & 0.1159 & 5.113 & 16.73 & 19 \\
 & STID & 0.1122 & 4.299 & 13.48 & 0.1419 & 5.941 & 20.92 & 0.1542 & 6.397 & 20.22 & 0.1166 & 5.149 & 16.74 & 20 \\
 & DLinear & 0.1171 & 4.483 & 13.48 & 0.1441 & 6.029 & \cellcolor[HTML]{F3F3FF}20.64 & 0.1545 & 6.403 & \cellcolor[HTML]{E9E9FE}19.78 & 0.1186 & 5.234 & \cellcolor[HTML]{DBDBFE}16.29 & 21 \\
 & TimeBase & 0.1222 & 4.674 & 13.68 & 0.1437 & 6.008 & \cellcolor[HTML]{E9E9FE}20.38 & 0.1553 & 6.437 & \cellcolor[HTML]{E6E6FE}19.67 & 0.1209 & 5.342 & \cellcolor[HTML]{CFCFFE}16.19 & 22 \\
 & SegRNN & 0.1116 & 4.275 & 13.60 & 0.1418 & 5.932 & 20.98 & 0.1533 & 6.360 & 20.48 & 0.1196 & 5.280 & 17.03 & 23 \\
 & RMLP & 0.1135 & 4.344 & 13.51 & 0.1431 & 5.986 & 20.91 & 0.1562 & 6.476 & 20.39 & 0.1176 & 5.185 & 16.67 & 25 \\
 & SparseTSF & 0.1134 & 4.342 & 13.60 & 0.1420 & 5.941 & 20.99 & 0.1553 & 6.442 & 20.53 & 0.1176 & 5.193 & 16.75 & 27 \\
\multirow{-10}{*}{\rotatebox[origin=c]{90}{Channel-Independent}} & PhaseFormer & 0.1151 & 4.411 & 13.69 & 0.1432 & 5.993 & 21.03 & 0.1529 & 6.339 & 20.26 & 0.1226 & 5.423 & 17.07 & 28 \\
\midrule
& PMDformer & \cellcolor[HTML]{CFCFFE}0.1077 & \cellcolor[HTML]{C2C2FD}4.122 & \cellcolor[HTML]{E9E9FE}13.28 & \cellcolor[HTML]{E6E6FE}0.1384 & \cellcolor[HTML]{E6E6FE}5.790 & 20.70 & \cellcolor[HTML]{E9E9FE}0.1486 & \cellcolor[HTML]{E9E9FE}6.157 & \cellcolor[HTML]{F9F9FF}20.01 & \cellcolor[HTML]{C2C2FD}0.1125 & \cellcolor[HTML]{C2C2FD}4.962 & \cellcolor[HTML]{F3F3FF}16.39 & 4 \\
 & TimeFilter & 0.1099 & \cellcolor[HTML]{F9F9FF}4.206 & \cellcolor[HTML]{F3F3FF}13.30 & \cellcolor[HTML]{F3F3FF}0.1390 & \cellcolor[HTML]{F3F3FF}5.812 & \cellcolor[HTML]{F9F9FF}20.67 & \cellcolor[HTML]{EDEDFF}0.1494 & \cellcolor[HTML]{EDEDFF}6.187 & \cellcolor[HTML]{F3F3FF}20.00 & \cellcolor[HTML]{E6E6FE}0.1137 & \cellcolor[HTML]{E6E6FE}5.012 & 16.42 & 7 \\
 & Crossformer & \cellcolor[HTML]{F9F9FF}0.1096 & \cellcolor[HTML]{F3F3FF}4.199 & 13.42 & \cellcolor[HTML]{F3F3FF}0.1390 & \cellcolor[HTML]{F9F9FF}5.816 & 20.76 & 0.1533 & 6.355 & 20.46 & \cellcolor[HTML]{E9E9FE}0.1141 & \cellcolor[HTML]{EDEDFF}5.039 & 16.59 & 12 \\
 & SOFTS & 0.1111 & 4.252 & 13.45 & 0.1408 & 5.888 & 20.84 & 0.1533 & 6.354 & 20.29 & 0.1159 & 5.112 & 16.76 & 13 \\
 & ModernTCN & 0.1117 & 4.279 & 13.42 & 0.1416 & 5.925 & 20.86 & \cellcolor[HTML]{F9F9FF}0.1512 & \cellcolor[HTML]{F9F9FF}6.269 & 20.11 & 0.1171 & 5.171 & 16.78 & 14 \\
 & TQNet & 0.1117 & 4.277 & 13.44 & 0.1413 & 5.910 & 20.83 & 0.1537 & 6.373 & 20.25 & 0.1161 & 5.121 & 16.66 & 14 \\
 & iTransformer & 0.1117 & 4.276 & 13.47 & 0.1410 & 5.899 & 20.88 & 0.1518 & 6.290 & 20.16 & 0.1169 & 5.159 & 16.79 & 16 \\
 & Leddam & 0.1117 & 4.280 & 13.47 & 0.1417 & 5.928 & 20.87 & 0.1534 & 6.361 & 20.30 & 0.1163 & 5.133 & 16.68 & 18 \\
 & CrossGNN & 0.1125 & 4.309 & 13.59 & 0.1417 & 5.926 & 21.00 & 0.1542 & 6.392 & 20.45 & 0.1170 & 5.162 & 16.81 & 24 \\
\multirow{-10}{*}{\rotatebox[origin=c]{90}{Multivariate}} & DUET & 0.1147 & 4.397 & 13.57 & 0.1437 & 6.011 & 20.97 & 0.1539 & 6.379 & 20.23 & 0.1193 & 5.267 & 16.86 & 26 \\
\midrule
& TiDE & \cellcolor[HTML]{9E9EFC}0.1043 & \cellcolor[HTML]{9E9EFC}3.996 & \cellcolor[HTML]{9E9EFC}12.43 & \cellcolor[HTML]{B2B2FD}0.1293 & \cellcolor[HTML]{B2B2FD}5.397 & \cellcolor[HTML]{B2B2FD}18.96 & \cellcolor[HTML]{9E9EFC}0.1282 & \cellcolor[HTML]{9E9EFC}5.305 & \cellcolor[HTML]{9E9EFC}17.29 & \cellcolor[HTML]{B2B2FD}0.1104 & \cellcolor[HTML]{B2B2FD}4.866 & \cellcolor[HTML]{9E9EFC}15.81 & 1 \\
 & CrossLinear & \cellcolor[HTML]{B2B2FD}0.1051 & \cellcolor[HTML]{B2B2FD}4.032 & \cellcolor[HTML]{B2B2FD}12.65 & \cellcolor[HTML]{9E9EFC}0.1269 & \cellcolor[HTML]{9E9EFC}5.296 & \cellcolor[HTML]{9E9EFC}18.74 & \cellcolor[HTML]{C2C2FD}0.1328 & \cellcolor[HTML]{C2C2FD}5.505 & \cellcolor[HTML]{C2C2FD}18.43 & \cellcolor[HTML]{9E9EFC}0.1100 & \cellcolor[HTML]{9E9EFC}4.854 & \cellcolor[HTML]{B2B2FD}15.87 & 2 \\
 & XLinear & \cellcolor[HTML]{C2C2FD}0.1075 & \cellcolor[HTML]{CFCFFE}4.124 & \cellcolor[HTML]{C2C2FD}12.86 & \cellcolor[HTML]{C2C2FD}0.1300 & \cellcolor[HTML]{C2C2FD}5.431 & \cellcolor[HTML]{C2C2FD}19.42 & \cellcolor[HTML]{B2B2FD}0.1326 & \cellcolor[HTML]{B2B2FD}5.493 & \cellcolor[HTML]{CFCFFE}18.52 & \cellcolor[HTML]{F3F3FF}0.1144 & \cellcolor[HTML]{F3F3FF}5.055 & \cellcolor[HTML]{E6E6FE}16.30 & 3 \\
 & CATS & 0.1103 & 4.225 & \cellcolor[HTML]{DBDBFE}12.95 & \cellcolor[HTML]{CFCFFE}0.1321 & \cellcolor[HTML]{CFCFFE}5.510 & \cellcolor[HTML]{CFCFFE}19.53 & \cellcolor[HTML]{DBDBFE}0.1379 & \cellcolor[HTML]{DBDBFE}5.703 & \cellcolor[HTML]{DBDBFE}18.76 & 0.1159 & 5.106 & \cellcolor[HTML]{C2C2FD}16.13 & 5 \\
 & DAG & \cellcolor[HTML]{EDEDFF}0.1090 & \cellcolor[HTML]{E9E9FE}4.174 & \cellcolor[HTML]{EDEDFF}13.30 & \cellcolor[HTML]{E9E9FE}0.1388 & \cellcolor[HTML]{E9E9FE}5.803 & \cellcolor[HTML]{EDEDFF}20.64 & \cellcolor[HTML]{E6E6FE}0.1476 & \cellcolor[HTML]{E6E6FE}6.115 & \cellcolor[HTML]{EDEDFF}19.91 & \cellcolor[HTML]{E9E9FE}0.1141 & \cellcolor[HTML]{E9E9FE}5.033 & 16.55 & 6 \\
\multirow{-6}{*}{\rotatebox[origin=c]{90}{Exogenous}} & TimeXer & \cellcolor[HTML]{F3F3FF}0.1093 & 4.207 & \cellcolor[HTML]{CFCFFE}12.93 & \cellcolor[HTML]{E9E9FE}0.1388 & \cellcolor[HTML]{EDEDFF}5.804 & \cellcolor[HTML]{E6E6FE}20.14 & \cellcolor[HTML]{CFCFFE}0.1371 & \cellcolor[HTML]{CFCFFE}5.687 & \cellcolor[HTML]{B2B2FD}18.42 & 0.1172 & 5.177 & \cellcolor[HTML]{F9F9FF}16.40 & 8 \\
\midrule
& Chronos-2 & \cellcolor[HTML]{E9E9FE}0.1089 & \cellcolor[HTML]{E9E9FE}4.174 & \cellcolor[HTML]{E6E6FE}13.24 & \cellcolor[HTML]{DBDBFE}0.1363 & \cellcolor[HTML]{DBDBFE}5.703 & \cellcolor[HTML]{DBDBFE}20.08 & 0.1523 & 6.319 & 20.32 & 0.1155 & 5.101 & \cellcolor[HTML]{EDEDFF}16.34 & 9 \\
 & Timer-S1 & 0.1121 & 4.295 & 13.48 & 0.1417 & 5.927 & 20.86 & 0.1539 & 6.384 & 20.37 & \cellcolor[HTML]{F9F9FF}0.1153 & \cellcolor[HTML]{F9F9FF}5.090 & 16.54 & 17 \\
 & TimesFM 2.5 & 0.1160 & 4.457 & 13.69 & 0.1448 & 6.062 & 21.04 & 0.1585 & 6.581 & 20.58 & 0.1208 & 5.341 & 16.88 & 30 \\
\multirow{-4}{*}{\rotatebox[origin=c]{90}{Foundation}} & Moirai 2.0 & 0.1218 & 4.678 & 13.80 & 0.1466 & 6.130 & 20.92 & 0.1589 & 6.593 & 20.38 & 0.1260 & 5.572 & 17.06 & 31 \\
\bottomrule
\end{tabular}
}
\label{tab:main_results}
\end{table*}

\begin{table*}[t]
\centering
\caption{Eight-week-ahead forecasting performance, reporting long-horizon point-wise errors, day-level trend quality (i.e.,
directional accuracy of day and correlation of day), and first-week accuracy. Rank is calculated as the arithmetic mean of 
metric ranks across all experimental settings.}
\resizebox{0.78\textwidth}{!}{
\begin{tabular}{c|c|ccc|c|cc|c|ccc|c}
\toprule
\multicolumn{1}{c|}{} & Setting & \multicolumn{4}{c|}{Overall} & \multicolumn{3}{c|}{Trend} & \multicolumn{4}{c}{First Week} \\
 \cmidrule(lr){3-6} \cmidrule(lr){7-9} \cmidrule(lr){10-13}
& Metrics & WMAPE & MAE & RMSE & Rank & Acc.($\uparrow$) & Corr.($\uparrow$) & Rank & \multicolumn{1}{c}{WMAPE} & \multicolumn{1}{c}{MAE} & \multicolumn{1}{c|}{RMSE} & \multicolumn{1}{c}{Rank} \\
 \midrule 
& SeasonalMean & 0.1485 & 5.661 & 14.94 & 28 & 0.5875 & 0.2059 & 28 & 0.1391 & 5.349 & 14.86 & 31 \\
\multirow{-2}{*}{\rotatebox[origin=c]{90}{Sim.}} & SeasonalNaive & 0.1731 & 6.582 & 19.63 & 31 & 0.5518 & 0.1246 & 30 & 0.1613 & 6.182 & 19.36 & 32 \\
\midrule
& PETformer & \cellcolor[HTML]{C2C2FD}0.1278 & \cellcolor[HTML]{C2C2FD}4.879 & \cellcolor[HTML]{E6E6FE}14.58 & 2 & 0.6046 & \cellcolor[HTML]{EDEDFF}0.2549 & 12 & \cellcolor[HTML]{B2B2FD}0.1162 & \cellcolor[HTML]{B2B2FD}4.463 & \cellcolor[HTML]{CFCFFE}14.33 & 2 \\
 & TimeMixer & \cellcolor[HTML]{E6E6FE}0.1294 & \cellcolor[HTML]{DBDBFE}4.940 & \cellcolor[HTML]{DBDBFE}14.57 & 3 & \cellcolor[HTML]{F9F9FF}0.6061 & \cellcolor[HTML]{E6E6FE}0.2679 & 9 & \cellcolor[HTML]{DBDBFE}0.1182 & \cellcolor[HTML]{DBDBFE}4.542 & \cellcolor[HTML]{F3F3FF}14.43 & 3 \\
 & SegRNN & \cellcolor[HTML]{B2B2FD}0.1273 & \cellcolor[HTML]{B2B2FD}4.862 & 14.66 & 4 & 0.6059 & 0.2470 & 13 & \cellcolor[HTML]{CFCFFE}0.1173 & \cellcolor[HTML]{CFCFFE}4.508 & 14.48 & 6 \\
 & PatchTST & \cellcolor[HTML]{EDEDFF}0.1299 & \cellcolor[HTML]{EDEDFF}4.957 & \cellcolor[HTML]{E9E9FE}14.58 & 7 & 0.6001 & 0.2457 & 14 & \cellcolor[HTML]{E6E6FE}0.1188 & \cellcolor[HTML]{E6E6FE}4.562 & \cellcolor[HTML]{E9E9FE}14.42 & 3 \\
 & PhaseFormer & 0.1309 & 4.998 & 14.72 & 14 & 0.5983 & 0.2370 & 19 & \cellcolor[HTML]{E9E9FE}0.1190 & \cellcolor[HTML]{E9E9FE}4.568 & 14.50 & 10 \\
 & SparseTSF & 0.1338 & 5.106 & 14.92 & 19 & 0.5956 & 0.2121 & 24 & \cellcolor[HTML]{F3F3FF}0.1191 & \cellcolor[HTML]{F3F3FF}4.572 & 14.54 & 11 \\
 & STID & 0.1370 & 5.233 & 14.97 & 23 & 0.6026 & 0.2343 & 17 & 0.1254 & 4.818 & 14.73 & 27 \\
 & TimeBase & 0.1385 & 5.279 & 14.74 & 22 & 0.5895 & 0.1949 & 26 & 0.1239 & 4.752 & 14.54 & 25 \\
 & RMLP & 0.1398 & 5.322 & 14.99 & 28 & 0.5919 & 0.2346 & 21 & 0.1266 & 4.856 & 14.65 & 27 \\
\multirow{-10}{*}{\rotatebox[origin=c]{90}{Channel-Independent}}  & DLinear & 0.1396 & 5.318 & 14.84 & 25 & 0.5889 & 0.2282 & 25 & 0.1272 & 4.876 & 14.67 & 29 \\
\midrule
& ModernTCN & \cellcolor[HTML]{9E9EFC}0.1262 & \cellcolor[HTML]{9E9EFC}4.816 & \cellcolor[HTML]{B2B2FD}14.34 & 1 & \cellcolor[HTML]{E6E6FE}0.6089 & \cellcolor[HTML]{B2B2FD}0.3273 & 3 & \cellcolor[HTML]{F9F9FF}0.1193 & 4.587 & \cellcolor[HTML]{C2C2FD}14.28 & 8 \\
 & Leddam & \cellcolor[HTML]{E9E9FE}0.1295 & \cellcolor[HTML]{E9E9FE}4.945 & \cellcolor[HTML]{C2C2FD}14.54 & 4 & \cellcolor[HTML]{B2B2FD}0.6110 & \cellcolor[HTML]{C2C2FD}0.2837 & 2 & 0.1194 & 4.586 & \cellcolor[HTML]{DBDBFE}14.38 & 9 \\
 & SOFTS & \cellcolor[HTML]{CFCFFE}0.1280 & \cellcolor[HTML]{CFCFFE}4.886 & \cellcolor[HTML]{F9F9FF}14.64 & 6 & \cellcolor[HTML]{E9E9FE}0.6078 & \cellcolor[HTML]{CFCFFE}0.2783 & 4 & \cellcolor[HTML]{E9E9FE}0.1190 & \cellcolor[HTML]{EDEDFF}4.572 & 14.59 & 13 \\
 & iTransformer & \cellcolor[HTML]{F3F3FF}0.1302 & \cellcolor[HTML]{F3F3FF}4.972 & 14.70 & 11 & \cellcolor[HTML]{CFCFFE}0.6094 & \cellcolor[HTML]{F3F3FF}0.2527 & 5 & 0.1207 & 4.637 & 14.56 & 20 \\
 & PMDformer & 0.1316 & 5.022 & \cellcolor[HTML]{EDEDFF}14.63 & 11 & 0.5959 & 0.2402 & 20 & 0.1202 & 4.613 & \cellcolor[HTML]{E6E6FE}14.40 & 11 \\
 & CrossGNN & 0.1323 & 5.054 & 14.70 & 15 & 0.5952 & 0.2222 & 23 & 0.1200 & 4.610 & 14.46 & 15 \\
 & TQNet & 0.1328 & 5.075 & 14.74 & 17 & 0.6019 & 0.2367 & 16 & 0.1213 & 4.659 & 14.53 & 20 \\
 & TimeFilter & 0.1340 & 5.112 & 14.66 & 17 & 0.5992 & 0.2439 & 15 & 0.1227 & 4.708 & 14.51 & 22 \\
 & DUET & 0.1338 & 5.122 & 14.89 & 20 & \cellcolor[HTML]{C2C2FD}0.6106 & 0.2476 & 6 & 0.1235 & 4.756 & 14.69 & 26 \\
\multirow{-10}{*}{\rotatebox[origin=c]{90}{Multivariate}} & Crossformer & 0.1348 & 5.187 & 15.21 & 24 & \cellcolor[HTML]{CFCFFE}0.6094 & 0.2476 & 7 & 0.1210 & 4.663 & 14.75 & 23 \\
\midrule
& TiDE & 0.1328 & 5.069 & \cellcolor[HTML]{9E9EFC}14.32 & 10 & \cellcolor[HTML]{9E9EFC}0.6137 & \cellcolor[HTML]{9E9EFC}0.3715 & 1 & 0.1194 & \cellcolor[HTML]{F9F9FF}4.582 & \cellcolor[HTML]{9E9EFC}13.99 & 7 \\
 & CrossLinear & \cellcolor[HTML]{DBDBFE}0.1293 & \cellcolor[HTML]{E6E6FE}4.941 & 14.66 & 7 & \cellcolor[HTML]{EDEDFF}0.6077 & \cellcolor[HTML]{E9E9FE}0.2671 & 7 & 0.1194 & 4.590 & 14.49 & 14 \\
 & TimeXer & 0.1317 & 5.040 & \cellcolor[HTML]{CFCFFE}14.56 & 9 & 0.6058 & \cellcolor[HTML]{DBDBFE}0.2707 & 10 & 0.1218 & 4.690 & \cellcolor[HTML]{EDEDFF}14.42 & 17 \\
 & XLinear & \cellcolor[HTML]{F9F9FF}0.1303 & \cellcolor[HTML]{F9F9FF}4.975 & 14.73 & 13 & \cellcolor[HTML]{F3F3FF}0.6072 & \cellcolor[HTML]{F9F9FF}0.2492 & 11 & 0.1211 & 4.658 & 14.53 & 19 \\
 & DAG & 0.1335 & 5.099 & \cellcolor[HTML]{F3F3FF}14.63 & 15 & 0.5975 & 0.2428 & 17 & 0.1222 & 4.695 & \cellcolor[HTML]{F9F9FF}14.45 & 18 \\
\multirow{-6}{*}{\rotatebox[origin=c]{90}{Exogenous}} & CATS & 0.1349 & 5.155 & 14.85 & 21 & 0.5961 & 0.2200 & 21 & 0.1230 & 4.726 & 14.56 & 24 \\
\midrule
& Chronos-2 & 0.1374 & 5.270 & 15.07 & 26 & 0.5905 & 0.1702 & 26 & \cellcolor[HTML]{9E9EFC}0.1140 & \cellcolor[HTML]{9E9EFC}4.382 & \cellcolor[HTML]{B2B2FD}14.27 & 1 \\
 & Timer-S1 & 0.1379 & 5.288 & 15.11 & 27 & 0.5583 & 0.0910 & 31 & \cellcolor[HTML]{C2C2FD}0.1168 & \cellcolor[HTML]{C2C2FD}4.487 & 14.49 & 3 \\
 & TimesFM 2.5 & 0.1564 & 6.026 & 15.66 & 30 & 0.5737 & 0.1128 & 29 & \cellcolor[HTML]{F9F9FF}0.1193 & 4.589 & 14.59 & 16 \\
\multirow{-4}{*}{\rotatebox[origin=c]{90}{Foundation}} & Moirai 2.0 & 0.2039 & 7.925 & 18.48 & 32 & 0.5405 & 0.0582 & 32 & 0.1263 & 4.864 & 14.79 & 30 \\
\bottomrule
\end{tabular}
}
\label{tab:long_horizon}
\end{table*}

\subsection{Implementation Details}

All experiments are implemented in Python 3.11 with PyTorch~\cite{pytorch} and conducted on a cluster of eight NVIDIA RTX A6000 GPUs.
Forecasting samples are generated with task-specific sliding-window strides: 48 steps, i.e., one day, for week-ahead forecasting, and 336 steps, i.e., one week, for 8-week-ahead forecasting.

Models are trained using the MAE loss and optimized with AdamW~\cite{adamw} under a unified hyperparameter search protocol.
For each method, we search the learning rate from $\{3\times10^{-5}, 1\times10^{-4}, 3\times10^{-4}, 1\times10^{-3}, 3\times10^{-3}\}$ and report the results under the best configuration.
Unless otherwise specified, we use a batch size of 128 and reduce it when out-of-memory issues occur.
Early stopping is applied if the validation performance does not improve for three consecutive epochs.

For models that support exogenous inputs, future-known exogenous variables are provided through their model-specific interfaces; otherwise, only historical endogenous variables are used.

\section{Experiments}
\label{sec:experiments}

\subsection{Benchmark Results}
\label{subsec:benchmark_results}

\subsubsection{Regular Forecasting}
\label{subsubsec:regular_forecasting}

Table~\ref{tab:main_results} reports the results of regular week-ahead forecasting.
Overall, exogenous-aware models achieve the strongest performance across both overall evaluation and the three scenario-specific evaluations: \textit{Weather Disturbance}, \textit{Holiday Effect}, and \textit{Large-scale Event Impact}.
Among them, TiDE~\cite{tide} achieves consistently strong performance across different metrics and exogenous scenarios, highlighting the importance of effectively using future-known covariates for modeling demand deviations caused by external disturbances.

Some endogenous-only models, including channel-independent models (e.g., PETformer~\cite{petformer}) and multivariate models (e.g., PMDformer~\cite{pmdformer}), remain competitive in overall WMAPE and MAE.
This reflects the strong daily and weekly periodicity of regular ride-hailing demand, which can be captured from historical target values alone.
However, without access to future-known exogenous variables, these models are less effective under exogenous scenarios and also show less favorable overall RMSE.
This indicates that historical endogenous patterns alone are insufficient for modeling high-magnitude or abrupt demand changes.

Time series foundation models also show promising zero-shot transferability.
For example, Chronos-2~\cite{chronos2} achieves performance comparable to several task-specific trained models, suggesting the potential of large-scale pretrained forecasting models.
Nevertheless, they still lag behind specialized exogenous-aware models, especially under exogenous scenarios where future-known external information is important.

In summary, these results show that future-known exogenous variables are valuable for regular week-ahead forecasting, improving both overall accuracy and scenario-specific performance under weather, holiday, and large-scale event impacts.

\subsubsection{Long-Horizon Forecasting}
\label{subsubsec:long_horizon_forecasting}

Table~\ref{tab:long_horizon} presents the results of long-horizon 8-week-ahead forecasting.
Different from week-ahead forecasting, this setting evaluates long-horizon pointwise accuracy, day-level trend quality, and first-week accuracy, reflecting fine-grained numerical precision, macro-level trajectory modeling, and near-term accuracy preservation, respectively.

The results show that exogenous-aware models do not maintain the clear advantage observed in regular week-ahead forecasting.
Instead, several endogenous-only models achieve stronger overall accuracy.
This suggests that long-horizon ride-hailing forecasting relies heavily on stable temporal structures, such as weekly seasonality and long-term demand evolution.
It also indicates that simply adding exogenous inputs does not guarantee better long-horizon performance, and that current exogenous-aware architectures may not fully exploit such information over extended horizons.

Another important observation is that pointwise accuracy and trend quality are not strictly aligned.
Models with lower WMAPE, MAE, or RMSE do not necessarily achieve the best Acc. or Corr.
This indicates that local numerical precision and macro-level trajectory modeling capture distinct forecasting capabilities, and that existing models still struggle to achieve both simultaneously.
For long-horizon operational planning, an effective model should go beyond minimizing pointwise errors and reliably capture the direction and overall shape of future demand changes.

Moreover, several channel-independent models and foundation models achieve strong first-week accuracy.
This suggests that models with strong temporal pattern modeling can preserve near-term precision even when evaluated under an 8-week horizon.
By contrast, the advantages of modeling cross-variable dependencies or exogenous interactions in week-ahead forecasting do not necessarily transfer to the long-horizon setting.

In summary, no existing model dominates all dimensions of long-horizon forecasting.
This reveals an open challenge: future long-horizon ride-hailing forecasting models should balance long-horizon pointwise accuracy, trend quality, and first-week accuracy.

\subsubsection{Accuracy--Efficiency Trade-off}
\label{subsubsec:accuracy_efficiency}

\begin{figure*}[t]
  \centering
  \includegraphics[width=0.48\linewidth]{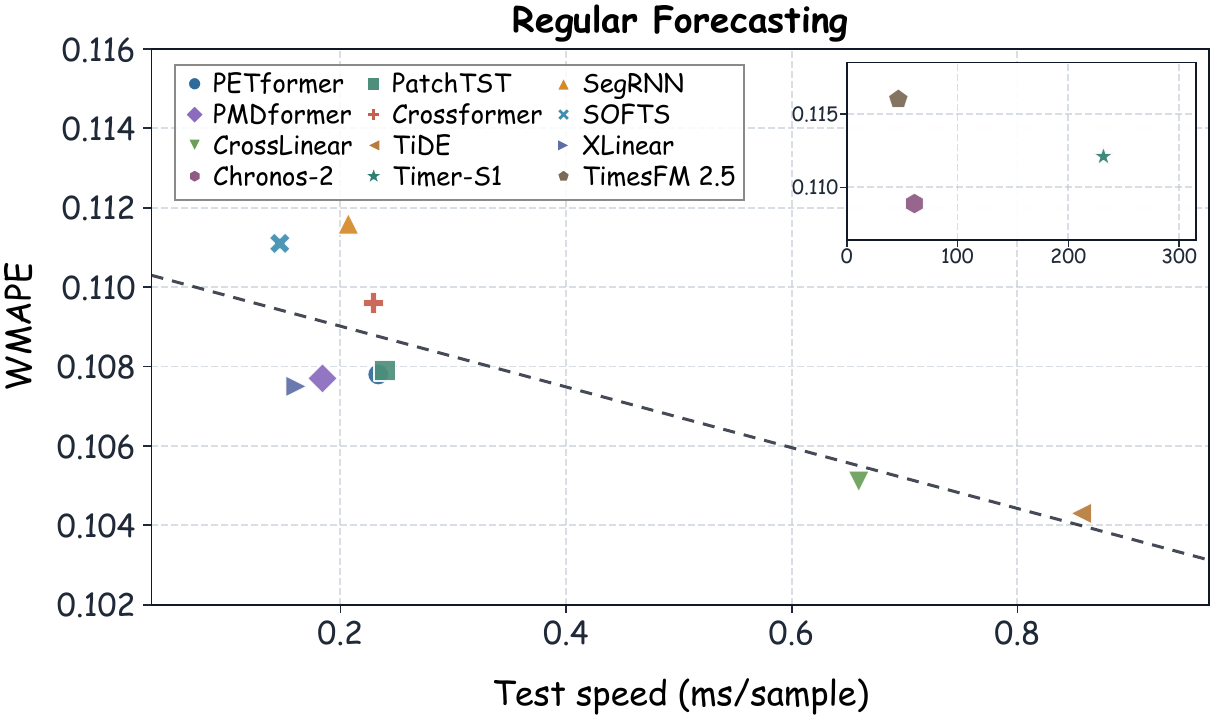}
  \hfill
  \includegraphics[width=0.48\linewidth]{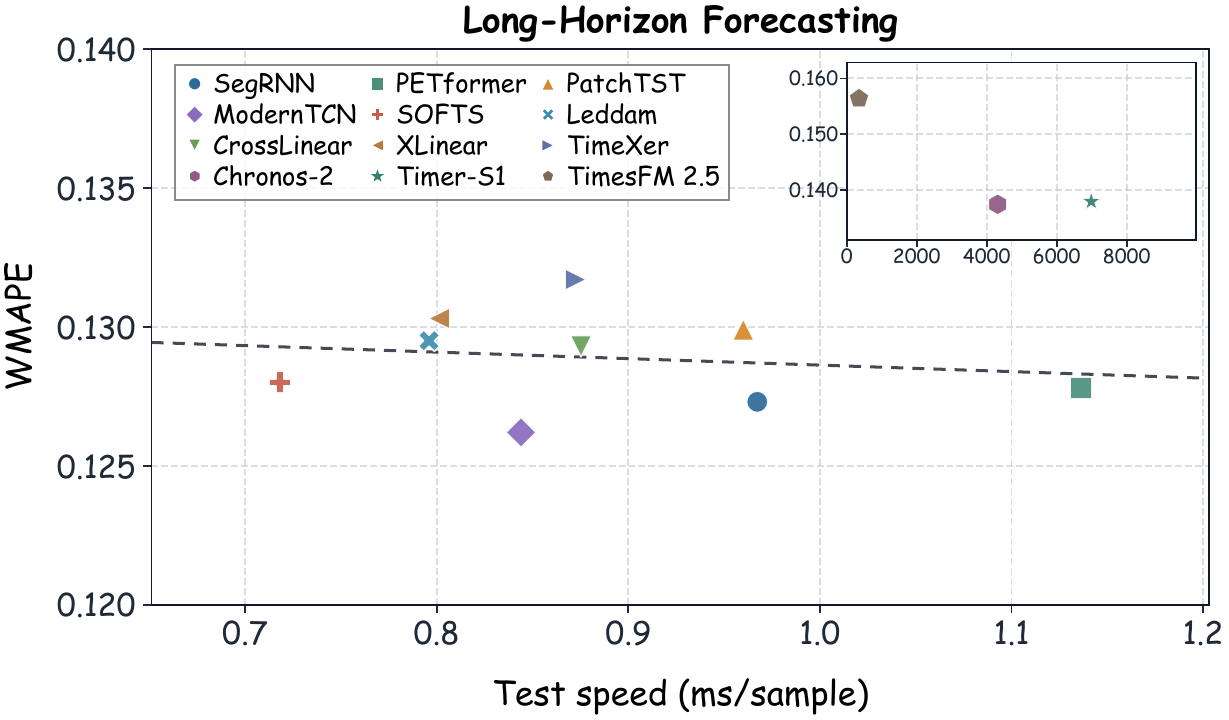}
  \caption{Efficiency of representative forecasting models under regular and long-horizon settings.}
  \label{fig:efficiency}
\end{figure*}

\begin{figure*}[t]
  \centering
  \includegraphics[width=0.33\linewidth]{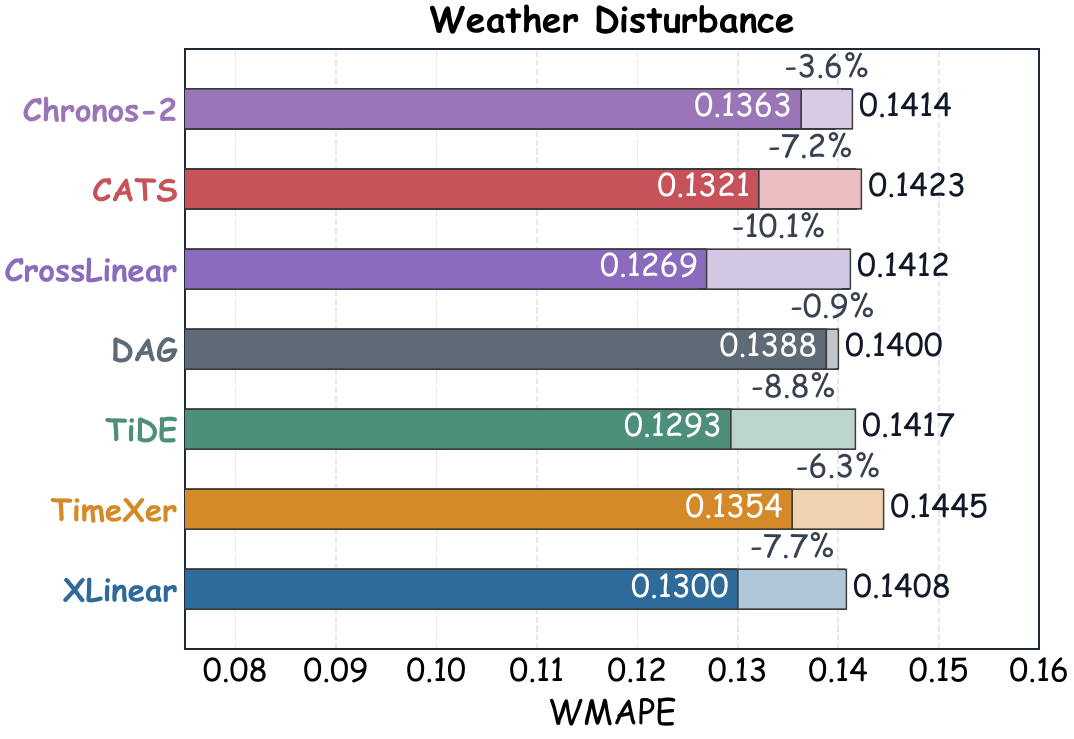}
  \hfill
  \includegraphics[width=0.33\linewidth]{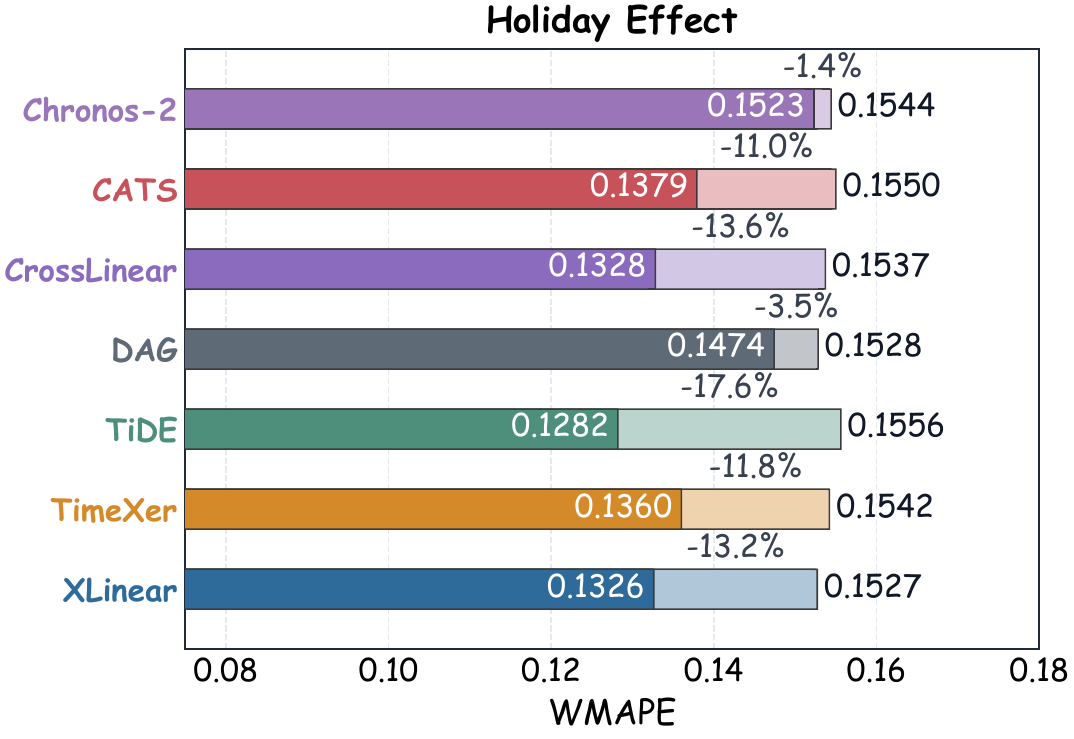}
  \hfill
  \includegraphics[width=0.33\linewidth]{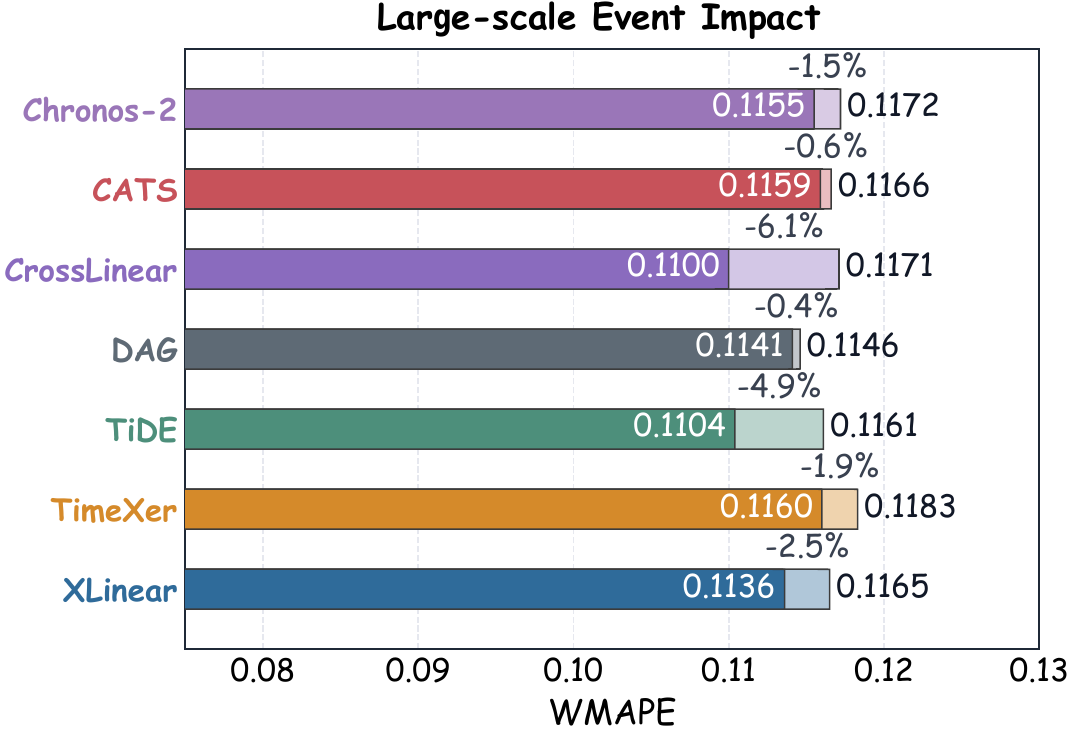}
  \caption{Effect of incorporating exogenous variables under three scenario-specific evaluations.}
  \label{fig:exo_ablation}
\end{figure*}

While Tables~\ref{tab:main_results} and~\ref{tab:long_horizon} focus on overall forecasting performance, computational efficiency is also critical for practical deployment.
Ride-hailing forecasting is typically performed over large-scale areas, multiple target variables, and repeated rolling prediction windows.
Thus, models with excessive inference latency may be difficult to deploy even if they achieve competitive accuracy.
To examine this trade-off, Figure~\ref{fig:efficiency} compares the forecasting accuracy and inference cost of representative models.

Figure~\ref{fig:efficiency} shows that task-specific deep models generally provide a more favorable accuracy--efficiency balance.
Their architectures are usually lightweight enough for efficient batched inference under the same benchmark protocol.
In contrast, time series foundation models tend to incur substantially higher inference cost due to their larger model scale and, in many cases, prediction paradigms that decompose the original multi-target task into multiple sub-tasks, such as independent univariate forecasts.

The trade-off also differs between regular and long-horizon forecasting.
In regular week-ahead forecasting, higher computational cost is more likely to bring accuracy gains, suggesting that larger capacity and better exogenous feature utilization can still be beneficial.
However, in 8-week-ahead forecasting, this relationship becomes much weaker.
This suggests that long-horizon performance is not simply constrained by model capacity, but also by the difficulty of capturing stable long-term trajectories.

These results suggest different research priorities for the two forecasting settings.
For regular week-ahead forecasting, improving pointwise accuracy remains valuable, as larger capacity and stronger exogenous-aware architectures may further improve the use of future-known covariates.
For long-horizon forecasting, however, simply increasing model complexity to reduce pointwise errors may be less effective.
Future work should place more emphasis on trend-aware modeling, including the accurate prediction of demand directions, trajectory shapes, and stable long-term temporal structures.

\subsection{Ablation and Analysis}
\label{subsec:ablation_analysis}
To better understand how different design factors affect forecasting performance, we conduct additional analyses under the regular week-ahead forecasting setting.

\begin{figure*}[t]
  \centering
  \includegraphics[width=0.33\linewidth]{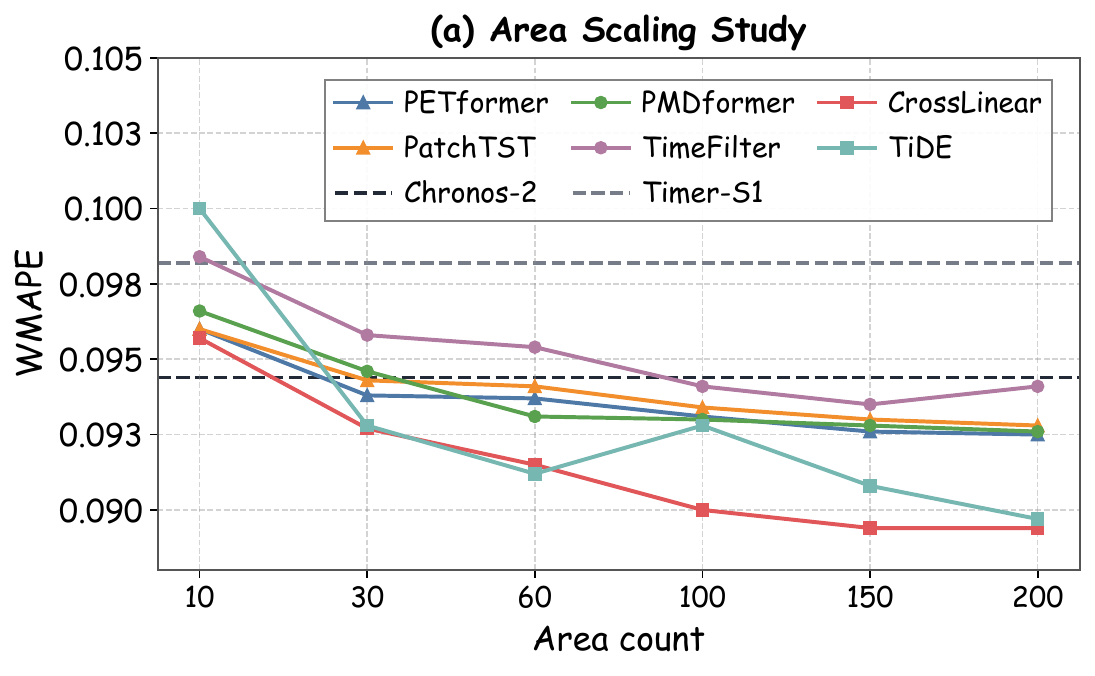}
  \hfill
  \includegraphics[width=0.33\linewidth]{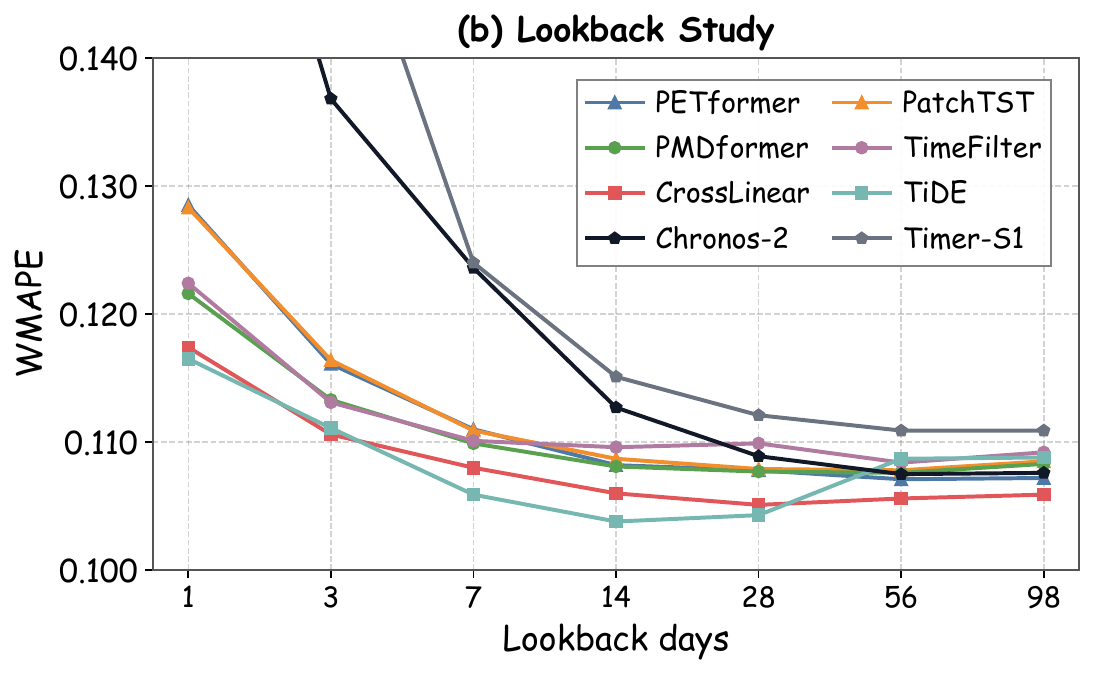}
  \hfill
  \includegraphics[width=0.33\linewidth]{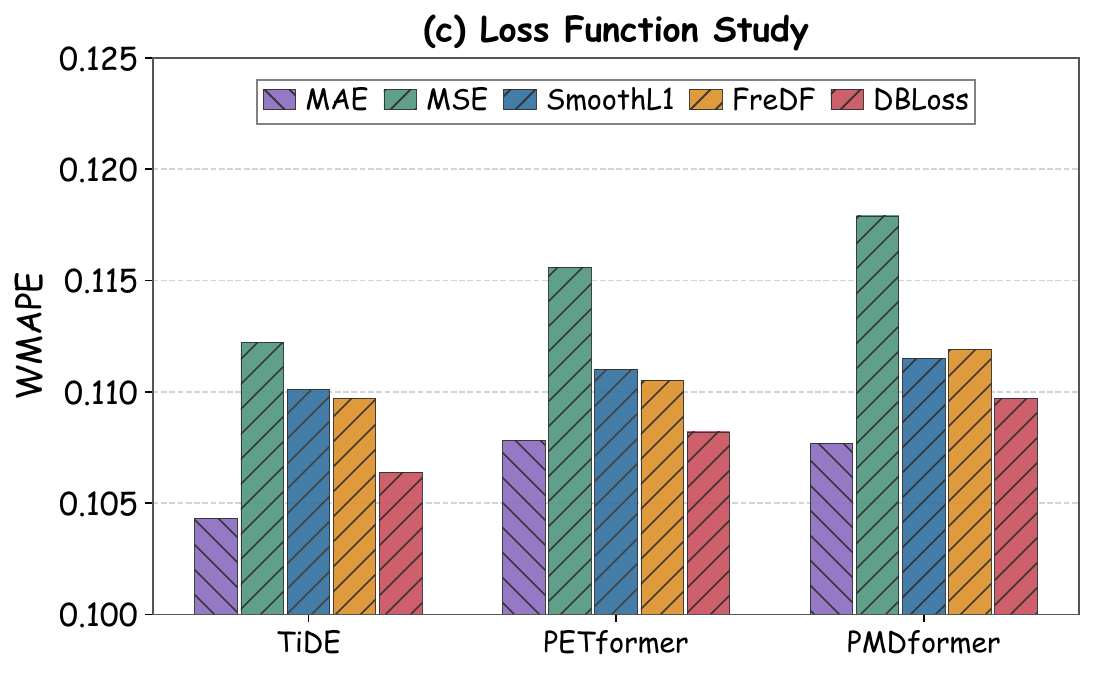}
  \caption{Studies on key benchmark design factors.
  (a) Effect of increasing the number of training areas while keeping the 10 target areas fixed.
  (b) Effect of varying the input lookback length.
  (c) Impact of different training losses.}
  \label{fig:study}
\end{figure*}

\subsubsection{Exogenous Variable Utilization}
\label{subsubsec:exogenous_utilization}

To further examine whether current models can benefit from future-known external information, we compare their performance with and without exogenous variables in Figure~\ref{fig:exo_ablation}.
The comparison is conducted under three scenario-specific evaluations.
Although Chronos-2~\cite{chronos2} is a zero-shot foundation model, it is also included because it natively supports exogenous inputs.

Figure~\ref{fig:exo_ablation} shows that most exogenous-aware models obtain clear improvements after incorporating external variables.
This suggests that the provided exogenous features contain useful predictive signals and that existing models can exploit them to varying degrees.
Among these methods, TiDE~\cite{tide} achieves the largest average improvement, with a relative gain of 10.4\%.

Chronos-2 also benefits from exogenous inputs under the zero-shot setting, suggesting that foundation models can exploit additional covariates to some extent.
However, its gains are smaller for categorical scenarios such as \textit{Holiday Effect} than for continuous disturbance scenarios such as \textit{Weather Disturbance}, since pretrained time series foundation models may not capture the downstream semantic mappings of discrete encodings~\cite{chronos2}.

The improvement under the \textit{Large-scale Event Impact} scenario is generally smaller than that under weather and holiday scenarios.
This reflects the difficulty of modeling sparse and heterogeneous event effects.
Unlike weather and holidays, large-scale events with explicit semantic meanings are often sparse across both areas and time, and exhaustive event collection is difficult in practice.
As a result, available event samples are limited, making it harder for models to learn reliable event-specific response patterns.

In summary, future-known exogenous variables provide clear benefits for ride-hailing forecasting, but their usefulness depends on variable type and scenario complexity.
Continuous disturbance signals such as weather are relatively easier to exploit, while categorical holidays and sparse large-scale events remain more challenging.
We further discuss these challenges in Section~\ref{sec:discussion}.

\subsubsection{Area Scaling Study}
\label{subsubsec:area_scaling}

Ride-Hailing contains parallel time series from 200 spatial areas, providing a natural setting to study whether forecasting models can benefit from larger-scale multi-area training.
This study focuses on a key question for global ride-hailing forecasting: \emph{when the target areas remain fixed, can adding more training areas help models learn transferable structures and improve area-specific forecasts?}

To answer this question, we fix 10 target areas for evaluation and gradually increase the number of training areas from 10 to 200.
The 10-area setting represents a limited-data regime where models can only learn from the target areas themselves, while larger settings introduce additional heterogeneous areas into training.
We evaluate representative top-performing models from each model family, and include zero-shot time series foundation models as reference baselines.

As shown in Figure~\ref{fig:study}(a), Chronos-2~\cite{chronos2} performs strongly when only 10 training areas are available, highlighting the advantage of pretrained foundation models under limited task-specific data.
However, as more areas are included in training, task-specific models continue to improve.
This indicates that the additional areas do not merely introduce heterogeneity or noise; instead, they provide useful transferable information that can improve forecasting on the fixed target areas.

This finding supports the global multi-area training design of RideBench.
Although different areas exhibit distinct demand patterns, they still share common temporal and exogenous structures, such as daily mobility cycles, weekly seasonality, and recurring responses to external factors.
A strong ride-hailing forecasting model should therefore learn from large-scale heterogeneous areas while producing reliable forecasts for each individual area.

The comparison with foundation models further reveals a data-scaling effect.
Zero-shot foundation models are competitive when task-specific data are scarce, but their advantage becomes less pronounced as more high-quality domain data are available for training.
This suggests that large-scale domain-specific data remain a key resource for building strong ride-hailing forecasting models, which is precisely the value of Ride-Hailing.

\subsubsection{Lookback Study}
\label{subsubsec:lookback_study}

The input lookback window determines how much historical context is available to the model, making it a key hyperparameter in time series forecasting~\cite{tqnet}.
We vary the input length to examine whether current models can benefit from longer temporal context.

As shown in Figure~\ref{fig:study}(b), most models improve as the lookback window increases.
This confirms that longer historical context provides useful information for ride-hailing forecasting, where daily, weekly, and longer-term patterns coexist.
It also suggests that well-designed forecasting models can exploit extended context to learn more robust temporal dependencies.

However, the benefit of longer lookback windows is not unlimited.
When the input length is extended to 98 days, several task-specific trained models show slight performance degradation.
This may be caused by two factors: excessively long contexts can introduce irrelevant or outdated information, and longer input windows reduce the number of available sliding-window training samples.
Thus, increasing the lookback length does not always lead to better forecasting.

Based on this trade-off, RideBench adopts a 28-day lookback window as the default setting.
This provides sufficient historical information to cover multiple weekly cycles, while keeping computational cost moderate and preserving enough training samples.

An additional observation is that time series foundation models are more sensitive to insufficient lookback length.
When the input window is short, their performance is noticeably weaker; only when the lookback reaches around 28 days do they approach the performance of task-specific trained models.
This is because zero-shot foundation models are not optimized on Ride-Hailing and must infer task-relevant temporal structures directly from the input context.
Therefore, adequate historical context is particularly important when applying foundation models to fine-grained ride-hailing forecasting.

\subsubsection{Loss Function Study}
\label{subsubsec:loss_study}

The training loss determines the optimization preference of forecasting models.
We compare commonly used objectives, including MSE, MAE, and Smooth L1~\cite{smoothl1loss}, as well as two time-series-specific losses, FreDF~\cite{fredf} and DBloss~\cite{dbloss}.
FreDF emphasizes frequency-domain discrepancies, while DBloss considers decomposed temporal components such as periodicity and trend.
These objectives are relevant to ride-hailing forecasting because the data exhibit strong daily and weekly periodicity.

As shown in Figure~\ref{fig:study}(c), MAE achieves the best overall performance.
This is consistent with the characteristics of ride-hailing demand.
Although the series contain regular periodic patterns, they are also affected by exogenous disturbances that can introduce abrupt peaks and large deviations.
In such settings, MSE may over-penalize a small number of large-error points and become overly sensitive to disturbance-induced outliers.
By contrast, MAE provides a more robust optimization objective under heavy-tailed forecasting errors.

FreDF and DBloss also show competitive performance compared with MSE and Smooth L1, suggesting that frequency- and decomposition-aware objectives can help capture periodic temporal structures.
However, they still underperform MAE in our setting.
This indicates that, for ride-hailing forecasting, robustness to exogenous disturbances is more critical than emphasizing periodic structures alone.
Therefore, we use MAE as the default training loss in RideBench.

\section{Discussion}
\label{sec:discussion}

\begin{figure*}[t]
  \centering
  \includegraphics[width=0.4\linewidth]{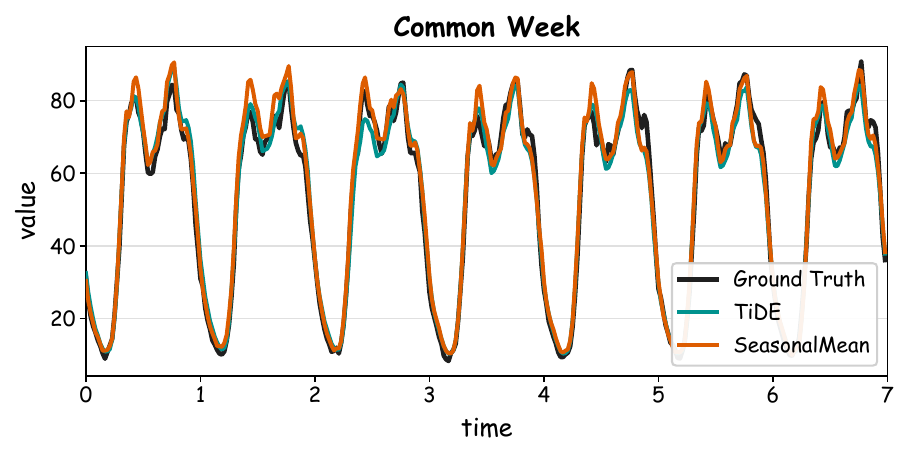}
  \centering
  \includegraphics[width=0.4\linewidth]{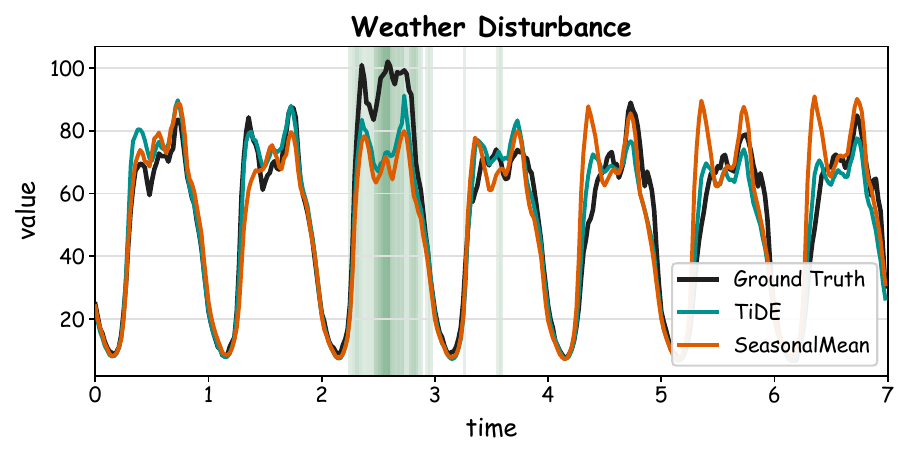}
  \centering
  \includegraphics[width=0.4\linewidth]{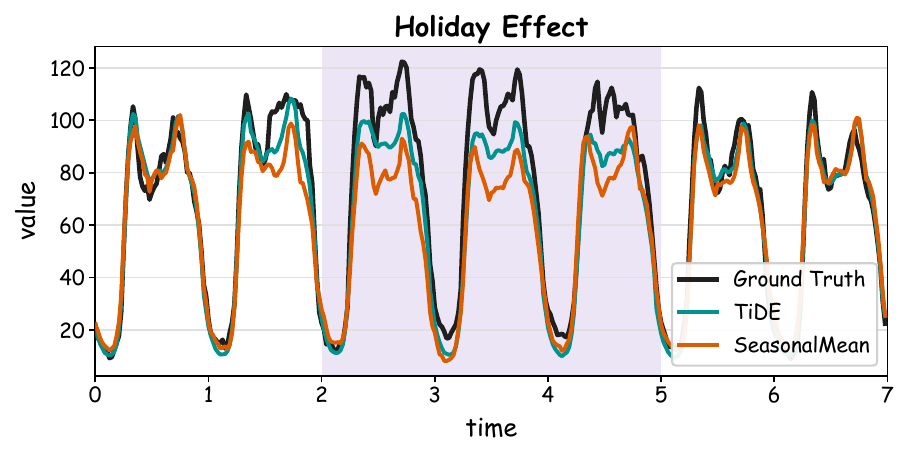}
  \centering
  \includegraphics[width=0.4\linewidth]{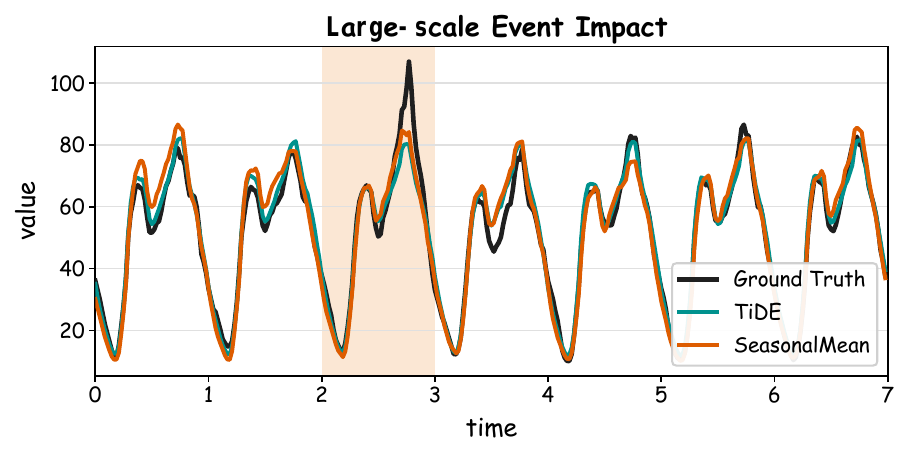}
  \caption{Forecasting visualization under different evaluation scenarios.}
  \label{fig:case_study}
\end{figure*}

By benchmarking representative deep learning forecasting methods developed in recent years, RideBench demonstrates the encouraging progress achieved by current models.
However, it also shows that existing methods still fall short of the requirements of  ride-hailing systems.
We summarize two key gaps below.

\textbf{Exogenous-driven pattern modeling remains insufficient.}
Although exogenous-aware models generally benefit from future-known external variables, they still cannot fully capture how external factors reshape ride-hailing dynamics.
As shown in Figure~\ref{fig:case_study}, even the best-performing exogenous-aware model, TiDE~\cite{tide}, only partially follows the ground-truth trajectory under strong external perturbations.
This suggests that current models can use exogenous variables as auxiliary signals, but still lack sufficient capability to model disturbance-induced pattern changes.
This limitation arises from both data side and model side.

On the data side, exogenous effects in ride-hailing systems are highly heterogeneous and temporally structured.
Weather impacts can be nonlinear or delayed; holidays involve pre-holiday, in-holiday, and post-holiday phases; and large-scale events are sparse, irregular, and temporally diverse.
Moreover, the same external factor may trigger different demand responses across areas due to differences in land use, commuting structure, supply-demand conditions, and event exposure.

On the model side, many existing exogenous-aware methods are not natively designed for the type of future-known categorical covariates considered in RideBench.
Models such as CrossLinear~\cite{crosslinear} and TimeXer~\cite{timexer} mainly focus on numerical covariates or historical contextual variables, whereas ride-hailing forecasting requires effective use of categorical holiday semantics, event types, and other future-known external indicators.
To make these methods applicable to RideBench, we extend their input interfaces to the prediction horizon and introduce categorical adapters such as one-hot encoding or learnable embeddings.
However, these adaptations are pragmatic compromises rather than principled solutions.
They allow models to access categorical variables, but do not necessarily enable them to understand their semantic structure or heterogeneous effects across areas.

\textbf{Long-horizon forecasting requires better multi-objective balance.}
RideBench further shows that no existing model dominates all aspects of 8-week-ahead forecasting, highlighting the difficulty of balancing multiple forecasting objectives.
As shown in Table~\ref{tab:long_horizon}, models with strong pointwise accuracy do not necessarily achieve the best trend quality, while models that preserve near-term accuracy may still fail to capture long-term trajectory changes.

However, this trade-off is important for practical ride-hailing deployment.
A desirable forecasting model should support both short-term operations and long-term planning: it should provide accurate near-term forecasts for operational decisions, reliable long-term trends for supply and resource planning, and stable performance over extended horizons.
Such a model would reduce the need to maintain separate forecasting systems for short-term dispatching and long-term planning.
However, existing benchmarks rarely evaluate these requirements jointly, which may partly explain why current models pay limited attention to this multi-objective balance.

Overall, RideBench suggests that  ride-hailing forecasting requires models that can jointly handle future-known exogenous variables, categorical event semantics, spatial heterogeneity, and long-horizon planning requirements.
We hope Ride-Hailing and RideBench can help the community develop more exogenous-aware, scalable, and deployment-oriented forecasting models.

\section{Conclusion}
\label{sec:conclusion}
In this paper, we introduced \textbf{Ride-Hailing}, a synthesized ride-hailing time series dataset, and built \textbf{RideBench} upon it as a unified benchmark for exogenous-aware ride-hailing forecasting under regular week-ahead and long-horizon 8-week-ahead settings, with standardized processing, global multi-area training, scenario-specific evaluation, and long-horizon trend evaluation.
Through evaluating more than 30 representative methods, we find that recent deep learning models have made encouraging progress, and that future-known exogenous variables bring clear benefits in regular week-ahead forecasting, especially under external disturbance scenarios.
Further analyses show the value of scalable multi-area training, sufficient historical context, and robust training objectives.
RideBench also highlights two remaining gaps for real-world deployment: current exogenous-aware models still have limited ability to capture disturbance-induced pattern changes, and long-horizon forecasting requires a better balance among pointwise accuracy, broad trend quality, and near-term reliability.
By providing a privacy-preserving synthesized dataset and unified evaluation protocols, we believe that RideBench can facilitate research on exogenous-aware, scalable, and deployment-oriented forecasting models for ride-hailing systems.

\begin{acks}
This work was supported by the CCF-DiDi GAIA Collaborative Research Funds (No. 202534) and the Shandong Provincial Natural Science Foundation Project (ZR2024LZH012).
\end{acks}


\bibliographystyle{ACM-Reference-Format}
\bibliography{reference}

\end{document}